\documentclass[final]{cvpr}

\usepackage{times}
\usepackage{epsfig}
\usepackage{graphicx}
\usepackage{amsmath}
\usepackage{amssymb}

\usepackage{pifont} 
\usepackage{multirow} 
\usepackage{booktabs} 

\usepackage[utf8]{inputenc} 
\usepackage{multirow}
\usepackage{siunitx}
\usepackage{tabularx}
\usepackage{booktabs}       
\usepackage{amsfonts}       
\usepackage{nicefrac}       
\usepackage{color}

\usepackage[toc,page]{appendix}

\usepackage[pagebackref=true,breaklinks=true,colorlinks,bookmarks=false]{hyperref}

\def\cvprPaperID{7155} 
\def\confYear{CVPR 2021}

\begin{document}

\title{Behavior-Driven Synthesis of Human Dynamics}

\author{
Andreas Blattmann\thanks{Indicates equal contribution.}  \qquad Timo Milbich\footnote[1]{} \qquad Michael Dorkenwald\footnote[1]{}  \qquad Bj{\"o}rn Ommer\\
Heidelberg Collaboratory for Image Processing, IWR \\
Heidelberg University, Germany\\
}

\maketitle

\begin{abstract}
Generating and representing human behavior are of major importance for various computer vision applications. Commonly, human video synthesis represents behavior as sequences of postures while directly predicting their likely progressions or merely changing the appearance of the depicted persons, thus not being able to exercise control over their actual behavior during the synthesis process. In contrast, controlled behavior synthesis and transfer across individuals requires a deep understanding of body dynamics and calls for a representation of behavior that is independent of appearance and also of specific postures. In this work, we present a model for human behavior synthesis which learns a dedicated representation of human dynamics independent of postures. Using this representation, we are able to change the behavior of a person depicted in an arbitrary posture, or to even directly transfer behavior observed in a given video sequence. To this end, we propose a conditional variational framework which explicitly disentangles posture from behavior. We demonstrate the effectiveness of our approach on this novel task, evaluating capturing, transferring, and sampling fine-grained, diverse behavior, both quantitatively and qualitatively. Project page is available at \small{\url{https://cutt.ly/5l7rXEp}}

\end{abstract}
\begin{figure*}[t]
\begin{center}
\includegraphics[width=.99\textwidth]{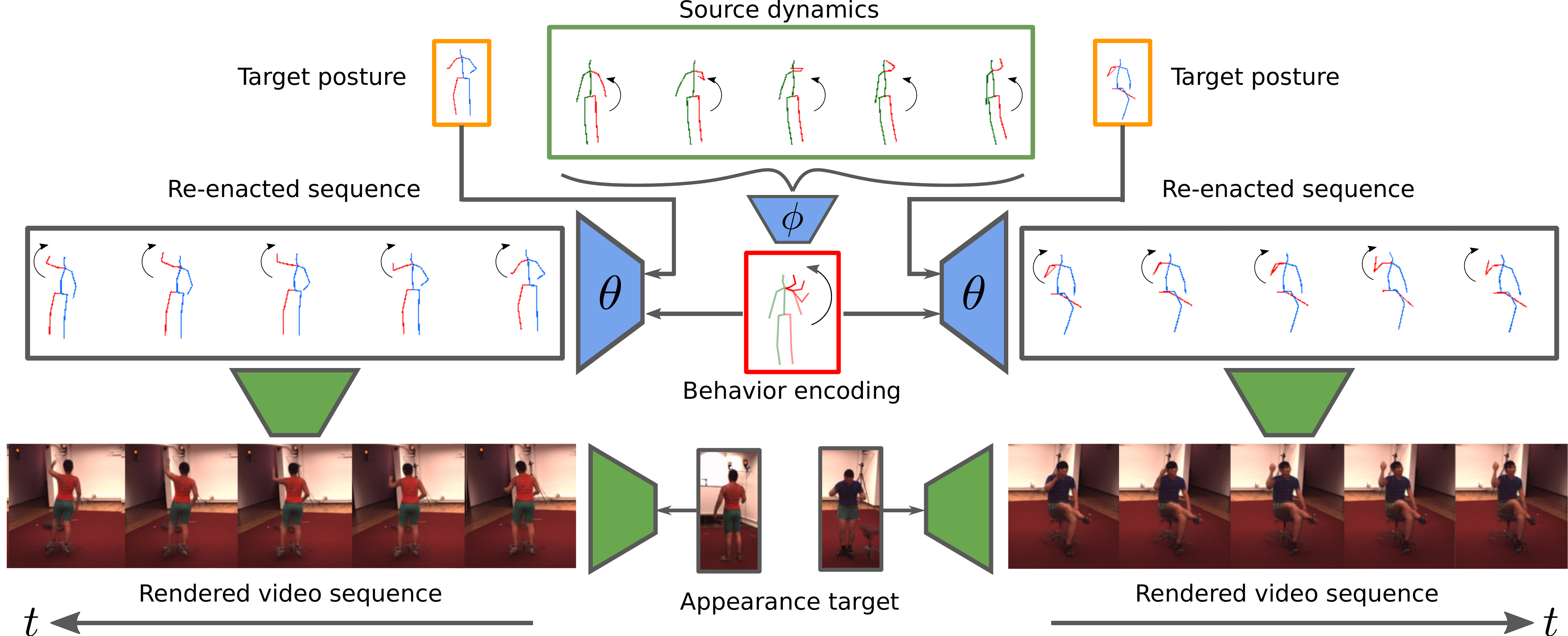}
\end{center}
   \caption{\textit{Our Approach for Behavior Transfer}. Given a source sequence of human dynamics our model infers a behavior encoding which is independent of posture. We can re-enact the behavior by combining it with an unrelated target posture and thus control the synthesis process. The resulting sequence is combined with an appearance to synthesize a video sequence.}
\label{fig:ffp}
\end{figure*}

\section{Introduction}
Understanding human appearance, posture and behavior are key problems of computer vision with numerous applications in autonomous driving~\cite{martin2019drive,masood2018detecting,gebert2019end}, surveillance~\cite{datta2002person,sultani2018real,xu2015learning}, medical treatment~\cite{uBAM_Biagio, wahl2017optogenetically} and beyond. While there has been major progress on representation~\cite{alphapose,densePose} and - with the advent of deep generative models~\cite{vae,gan} - synthesis~\cite{biggan,prog_grow} and manipulation~\cite{vunet,app_shape_dis_2, magnification_dorkenwald} of \textit{posture and appearance}, the understanding of representation and synthesis of \textit{behavior} is an open problem. 
\\
Human motor behavior is defined by the distinct dynamics of our limbs and the entire body. Take for example a person raising their arm. This is fully determined by the upward movement of the arm. Since the remaining body posture is mostly unaffected, the behavior can be directly performed independently of a particular initial body configuration such as a sitting or standing posture (cf. Fig. \ref{fig:ffp}). Moreover, rather complex behavior like running involves an interplay between certain body limbs, e.g. arms swinging synchronously with the movement of legs, and, thus, is naturally limited to certain postures to start with. To nevertheless enact such behavior from arbitrary starting poses, first a transition to fitting initial body configurations may be required - for instance, a sitting person needs to stand up before being able to walk. Finally, specific body features like size or build do not affect the ability to perform a walking behavior. 
\\
While behavior is eventually instantiated as a sequence of individual postures that can be observed in a video, this would be a suboptimal representation: We want the overall behavior to be the same, e.g. raising arm or walking, regardless of the initial posture it starts with. Although we are looking at different realizations it should still be represented as being the same behavior. Consequently, understanding, controlling, and synthesizing behavior calls for separate disentangled representations of the characteristic behavior and of individual (in particular the initial) posture.
In contrast, present work on human motion synthesis typically represents behavior directly by means of the observed sequence of postures~\cite{HPgan,martinez,yuan2020dlow,vid2vid}. Thus, as no explicit understanding and representation of behavior is developed, synthesizing human behavior has been limited to only changing person appearance ~\cite{vid2vid,dance_vid2vid,fewshot_vid2vid} or forecasting the most likely continuation of the depicted posture sequence~\cite{HPgan,martinez,yuan2020dlow,ActionAgnosticHP}. However, controlling such sequences, e.g. to re-enact a novel behavior by an observed person, asks for a posture independent representation which captures only the behavior dynamics to be transferred. Moreover, instantiating the re-enacted behavior requires combining these dynamics with the, potentially significantly different, posture of the target person.
\\
In this paper we propose a conditional variational generative model for controlled human behavior synthesis which only requires a collection of sequences without any class labels provided. Our models learns to understand the characteristic motor dynamics of behavior, which enables us to transfer behavior between videos. We learn a dedicated representation extracting these dynamics from pose sequences while factorizing out posture information. To this end, we propose an explicit disentanglement framework for behavior and posture based on an alternating optimization procedure while simultaneously controlling the information flow through our model. In particular, the explicit disentanglement allows our model to re-enact extracted behavior from arbitrary target postures and, if needed, to infer required corresponding transitions itself. Our experiments demonstrate qualitatively and quantitatively that our model meaningfully transfers behavior between sequences and is also able to sample novel and diverse behavior. Quantitative comparison against current approaches for human motion synthesis confirms the competitive performance of our approach.

\begin{figure*}
\begin{minipage}[l]{.5\textwidth}
    \centering
    \includegraphics[width=\textwidth]{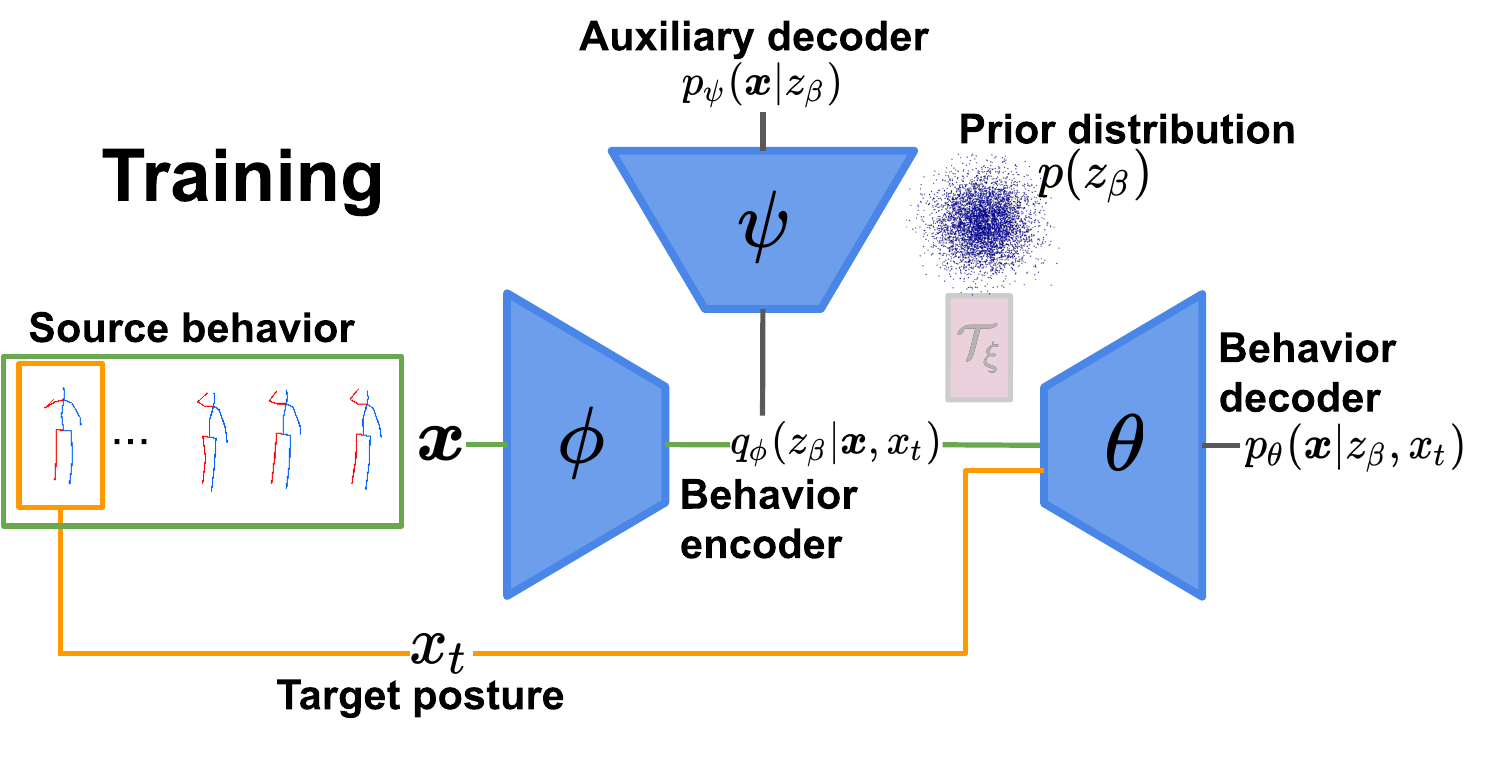}
    (a)
\end{minipage}
\hfill
\begin{minipage}[c]{0.5\textwidth}
    \centering
    \includegraphics[width=\textwidth]{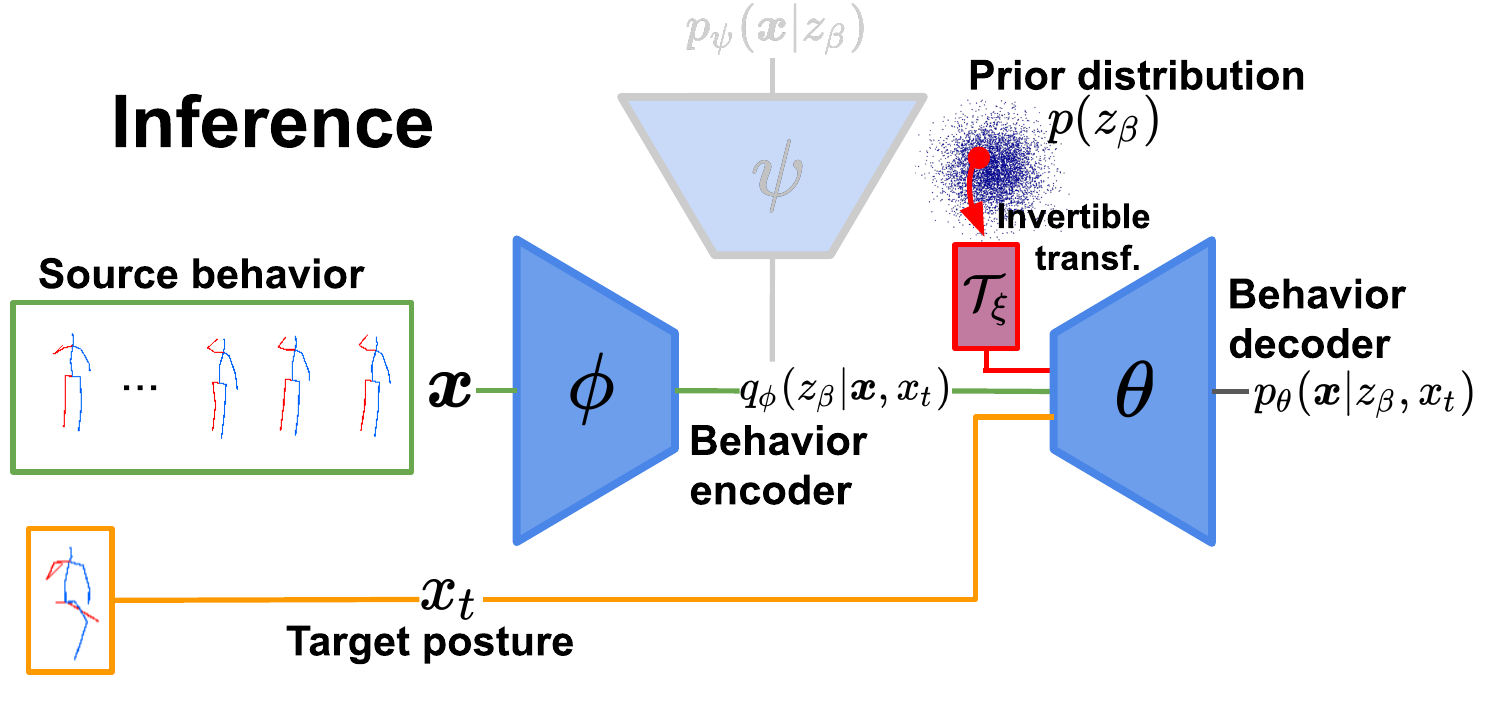}
(b)
\end{minipage}
\caption{\textit{Overview over model training (a) and inference (b)}. Each distribution is realized by a deep neural network. During training, the first posture of $\boldsymbol{x}$ serves as conditioning $x_t$ (yellow). Note, that consequently $x_t$ is also part of the encoder input since we do not have multiple training sequences $\boldsymbol{x}$ starting from the same posture $x_t$ available. In inference, i.e. after disentangling posture and behavior, we transfer source behavior (green) to an arbitrary target posture (yellow) or synthesize novel behavior from the prior distribution which is matched to $q_\phi(z_\beta|\boldsymbol{x},x_t)$ by a learned invertible transformation $\mathcal{T}_\xi$ (red).}
\label{fig:model}
\end{figure*}

\section{Related Work}
\noindent
\textbf{Static person rendering.}
Much work has been proposed to alter certain characteristics of humans depicted in static images like age, gender or body features~\cite{stylegan,glow,aegan,deep_feat_interpol} or synthesizing individual persons in different, unseen poses~\cite{vunet,esser2019unsupervised,app_shape_dis_1,app_shape_dis_2}. The latter task typically requires for explicit disentanglement between certain factors of interest, often depending on paired image data~\cite{esser2019unsupervised,app_shape_dis_1,cycleVAE}. While these approaches work well factors in static images, our work aims at transferring human behavior and, thus, requires disentanglement of a temporal factor, which is is significantly more complex.
\\
\textbf{Human video synthesis.}
Human video synthesis has been addressed in multiple ways. Some approaches synthesize videos directly in the pixel space~\cite{vidsynthVondrick,recycleGAN}. Due to the vast complexity of this problem, most approaches are based on mid-level representation of human shape, such as segmentation masks~\cite{vid2vid,game_vid2vid} or pose estimates~\cite{fewshot_vid2vid,dance_vid2vid,poseGuidedVidGen,hbugen,unsup_partbased}. Chan et al.~\cite{dance_vid2vid} generate video sequences of dancing persons by first learning correspondences between frames and postures before adding appearance information. A similar sequence-to-sequence translation task is performed in \cite{vid2vid,fewshot_vid2vid,lwb2019}. These works represent behavior directly on instantiated pose sequences, thus lacking the ability to exercise control. Our model understands and explicitly learns a behavior representation which can be used to transfer characteristic behavior dynamics between persons. Another line of research is future human motion prediction based on an initially observed posture sequence~\cite{unsup_ffp,struct_RNN,Fragkiadaki2015RecurrentNM,theposeknows}. Yuan et al.~\cite{kitani_dpp,yuan2020dlow} extend the future motion prediction task using multiple transformations on the latent space to increase the diversity of predicted motions. Chiu et al.~\cite{ActionAgnosticHP} propose a hierarchical multi-scale RNN to learn dependencies between individual postures. Martinez et al.~\cite{martinez} use residual RNN architectures to directly model motion velocities. Milbich et al.~\cite{iccv17_unsup_video} synthesize behavior by arranging frames from different video sequences based on nearest neighbour retrieval in a dedicated activity space. In contrast to our approach, these methods can not control the predicted behavior but only extrapolate the observed posture sequence. 
\\
\textbf{Controlled behavior synthesis.}
Controlling the behavior to be generated requires a considerable higher degree of understanding than unconditioned prediction or sequence translation. Recent works control mostly only in form of a small, fixed set of predefined actions~\cite{game_vid2vid,controllableVidGen}. Yang et al.~\cite{poseGuidedVidGen} condition the synthesis process on action labels. In contrast, we require only a collection of unpaired video sequences and condition the synthesis process on a dedicated representation of behavior independent of posture. DLow~\cite{yuan2020dlow} splits posture into different sets of keypoints to vary the diversity of predicted future movements for predefined body parts while keeping the others close to the groundtruth future sequence and, thus, cannot exercise detailed control. MT-VAE~\cite{MT-VAE_YAN} uses latent space arithmetics to enable transformations between different motions. However, transfer of more complex behavior is limited since only linear arithmetics are considered which makes a strong assumption on the latent space that typically cannot be met. In contrast, we learn a dedicated representation of behavior disentangled from posture. Hence our model naturally allows for recombination of behavior and posture.
\\
\textbf{Action recognition}
Action recognition \cite{carreira2018quo, wu2018compressed} aims at classifying a predefined set of actions from a given video, potentially based on intermediate representations such as 3D keypoints. Although our learned behavior representation is also based on keypoints, we aim at capturing behavior dynamics for detailed synthesis of full behavior. 
In contrast, action recognition learns discriminative representations which only focus on separating between action classes \cite{tishby2015deep}.


\section{Approach}
\label{sec:approach}
Our goal is to control and synthesize videos of human behavior. Since powerful pose estimators~\cite{alphapose,densePose} are readily available, pose sequences $\boldsymbol{x} = [x_0,...,x_n], x_i \in \mathbb{R}^{K \times 3}$ are directly used as a basis to represent the behavior observed in video~\cite{vid2vid,fewshot_vid2vid,dance_vid2vid}, e.g. for changing the depicted person's appearance or predicting likely sequence continuations. While this representation is sufficient to perform the aforementioned tasks, changing the posture sequence to re-enact a different behavior asks for a deeper understanding. Thus, behavior transfer requires separate representations modelling the characteristic motor dynamics of behavior and individual postures. We now present a generative model which extracts and represents human behavior $\beta$ from a source sequence $\boldsymbol{x}$ independent of the instantiated postures. Given an observed target posture $x_t$, e.g. in another video, we can then synthesize a re-enacted posture sequence and subsequently translate it to the video domain.
\\
Extracting behavior $\beta$ from $\boldsymbol{x}$ into a representation $z_\beta \in \mathbb{R}^D$ and subsequently re-combining it with a target pose $x_t$ to instantiate the behavior can be naturally formulated by means of latent variable models such as encoder-decoder frameworks. Such frameworks have been successfully applied for predicting future postures based on $\boldsymbol{x}$, i.e. directly extrapolating the observed posture sequence~\cite{yuan2020dlow,theposeknows}. However, as we seek to control the behavior to be generated, we require the latent representation $z_\beta$ to be disentangled from posture information.

\subsection{Synthesis using conditional generative models}
Generative models are powerful frameworks which are particularly suited for synthesis tasks. As we not only aim to learn a representation for behavior, but also need to extract it from our input sequences $\boldsymbol{x}$, variational autoencoders (VAE)~\cite{vae} are a natural choice. Such models approximate the true data distribution $p(\boldsymbol{x},z_\beta)$ which is assumed to follow the generative process $p(\boldsymbol{x},z_\beta) = p(\boldsymbol{x}|z_\beta)p(z_\beta)$. To optimize the intractable marginal log-likelihood $\mathbb{E}_{p(\boldsymbol{x})}[\log p_\theta(\boldsymbol{x})]$ of the model distribution $p_\theta(\boldsymbol{x}, z_\beta)$, a variational posterior $q_\phi(z_\beta|\boldsymbol{x})$ is introduced allowing to maximize a lower bound $L(p_\theta,q_\phi) \leq \mathbb{E}_{p(\boldsymbol{x})}[\log p(\boldsymbol{x})]$~\cite{vae}. Now, since we want to transfer behavior and condition it on arbitrary target postures, we condition the generative process~\cite{cvae} additionally on $x_t$, which modifies the lower variational bound and its optimization to
\begin{equation}
\begin{aligned}
    \max_{\theta,\phi} \; L(p_\theta,q_\phi) := \mathbb{E}_{q_\phi(z_\beta|\boldsymbol{x},x_t)} \left[ \log p_\theta(\boldsymbol{x}|z_\beta,x_t) \right] \\ - D_{\text{KL}}(q_\phi(z_\beta|\boldsymbol{x},x_t)||p(z_\beta))
    \label{eq:elbo}
\end{aligned}
\end{equation}
\noindent
where $p(z_\beta)$ is the prior on the latent representation $z_\beta$ which is typically modelled as a standard Gaussian distribution $\mathcal{N}(0,I)$. The first term of (\ref{eq:elbo}) can be considered to optimize the synthesis quality of the generator $p_\theta(\boldsymbol{x}|z_\beta,x_t)$ while the second part regularizes the encoder $q_\phi(z_\beta|\boldsymbol{x},x_t)$ to match the Gaussian prior.\\
Although our generator $p_\theta$ has access to both $z_\beta$ and the conditioning posture $x_t$, optimizing (\ref{eq:elbo}) will in general not encourage our model to learn a factorization of posture information and the behavior representation $z_\beta$. Moreover, we have no ground-truth provided for different behaviors starting from the same target posture $x_t$. Thus, we are only able to train our model by choosing $x_t$ to be the first posture of $\boldsymbol{x}$, which aggravates the need for an explicit disentanglement during the optimization process. 

\begin{figure*}[t]
\includegraphics[width=1\linewidth]{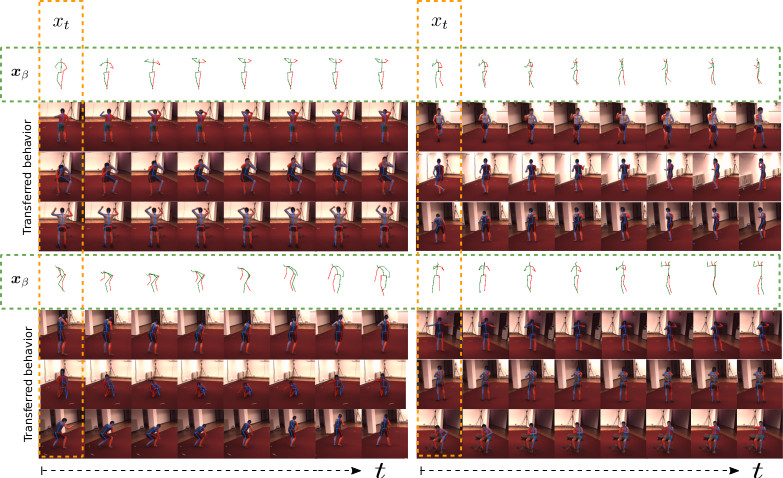}
   \caption{\textit{Behavior Transfer on Human3.6m}. We transfer fine-grained, characteristic body dynamics of an observed behavior $x_{\beta}$ to unrelated, significantly different target postures $x_t$. If required, the target posture is first adjusted by a transition phase before re-enacting the inferred behavior (e.g. top-right example, third row: walking starting from a bent down posture). Note that both transferred postures and images are generated by our models.}
\label{fig:behavior_transfer}
\vspace{-3mm}
\end{figure*}

\subsection{Disentangling posture from behavior}
While explicit disentanglement between factors of variation has been studied in the domain of static images~\cite{mathieu16,cycleVAE,app_shape_dis_1}, disentangling complex temporal information, however, is significantly still lacking. Existing works for static images typically resort to supervision by exploiting pairs of data samples sharing one factor while differing in the remaining factors~\cite{mathieu16,cycleVAE}, which allows for a natural disentanglement signal. Without having similar supervision available, we need to explicitly disentangle the posture information in $\boldsymbol{x}$ from our latent behavior representation $z_\beta$. To this end, we would ideally want to minimize the predictability of the individual postures in $\boldsymbol{x}$ given $z_\beta$. However, performing this operation directly on basis of our generator $p_\theta$ does not prevent the erasure of body dynamics as well. Instead, we can frame this task using an auxiliary generative model.  \\
Let $\hat{p}_\psi(\boldsymbol{x}|z_\beta)$ be a second generative model aiming at generating $\boldsymbol{x}$ from our behavior representation $z_\beta$ only, i.e. optimizing the log-likelihood,
\begin{equation}
    \max_{\psi} \; \mathbb{E}_{q_\phi(z_\beta|\boldsymbol{x},x_t)} \left[\log \hat{p}_\psi(\boldsymbol{x}|z_\beta) \right] \;.
    \label{eq:adv_likelihood}
\end{equation}
\noindent
Solving this task requires $\hat{p}_\psi(\boldsymbol{x}|z_\beta)$ to represent posture information which it has to be able to extract from $z_\beta$. Exploiting this, we can formulate our disentanglement task as an alternating optimization between our behavior model, i.e. $p_\theta, q_\phi$, optimizing $L(p_\theta, q_\phi)$ and $\hat{p}_\psi(\boldsymbol{x}|z_\beta)$ optimizing (\ref{eq:adv_likelihood}), both depending on the posterior $q_\phi(z_\beta|\boldsymbol{x},x_t)$.\footnote{However, note that (\ref{eq:adv_likelihood}) is not optimized over parameters $\phi$ and consequently does not affect $q_\phi(z_\beta|\boldsymbol{x},x_t)$.} To limit the predictability of $\hat{p}_\psi(\boldsymbol{x}|z_\beta)$, we extend (\ref{eq:elbo}) resulting in
\begin{equation}
    \max_{\theta, \phi} \; L(p_\theta, q_\phi) - \mathbb{E}_{q_\phi(z_\beta|\boldsymbol{x},x_t)} \left[\log \hat{p}_\psi(\boldsymbol{x}|z_\beta) \right] \;.
    \label{eq:constr_elbo}
\end{equation}
\noindent
This objective does not explicitly optimize parameters $\psi$, thus the predictability of $\hat{p}(\psi)$ can only be diminished by removing information about $\boldsymbol{x}$ from $z_\beta$. Further, note that $p_\theta$ has access to the conditional $x_t$ providing posture information and consequently only requires $q_\phi$ to provide missing dynamics to generate $\boldsymbol{x}$. The overall procedure can be considered as an adversarial task, alternating between optimizing (\ref{eq:adv_likelihood}) and (\ref{eq:constr_elbo}) in each training iteration. As a result, factoring out posture information from $z_\beta$ is indeed the most viable solution. Moreover, since posture information is excluded from our representation $z_\beta$, $p_\theta(\boldsymbol{x}|z_\beta,x_t)$ is required to infer a meaningful continuation of $x_t$ depicting behavior $\beta$.
\\
Due to the additional constraint in (\ref{eq:constr_elbo}), the already existing pressure to reduce the overall encoded information in $z_\beta$ imposed by $D_{\text{KL}}(q_\phi(z_\beta|\boldsymbol{x},x_t)||p(z_\beta))$ is further amplified. This also increases the risk of posterior collapses when using recurrent decoders~\cite{bowman16}, thus strongly affecting the generative process. Next, we discuss how to alleviate this problem by relaxing the information bottleneck.

\subsection{Relaxing the information bottleneck for improved synthesis}
The quality of synthesis depends on the expressiveness of $p_\theta(\boldsymbol{x}|z_\beta,x_t)$ which stands in contrast to the regularization of the variational posterior $q_\phi(z_\beta|\boldsymbol{x},x_t)$ in vanilla variational autoencoding settings~\cite{lossyVAE,infoVAE}. This becomes evident as the regularization $D_\text{KL}(q_\phi(z_\beta|\boldsymbol{x}, x_t)||p(z_\beta))$ minimizes an upper bound on the mutual information $I_{q_\phi}(\boldsymbol{x};z_\beta)$~\cite{rezaabad20}, thus reducing the information captured in $z_\beta$. Consequently, a typical solution is to explicitly maximize the mutual information~\cite{infoVAE,mutualAE}. However, computing reliable estimates of $I_{q_\phi}(\boldsymbol{x};z_\beta)$ is difficult for complex data~\cite{esser2019unsupervised,mine}. Instead, we resort to a relaxation of the regularization in the original variational problem by only optimizing $D_\text{KL}(q_\phi(z_\beta|\boldsymbol{x}, x_t)||p(z_\beta))$ to maintain a certain information budget $I_{\text{KL}}$, i.e. optimizing
\begin{equation}
\begin{aligned}
    \max_{\theta,\phi} \; & \mathbb{E}_{q_\phi(z_\beta|\boldsymbol{x},x_t)} \left[ \log p_\theta(\boldsymbol{x}|z_\beta,x_t) \right] \\ 
    & \text{s.t.} \; \; \; D_{\text{KL}}(q_\phi(z_\beta|\boldsymbol{x},x_t)||p(z_\beta)) \leq I_{\text{KL}} \;.
    \label{eq:info_elbo}
\end{aligned}
\end{equation}
\noindent
Similar to Peng et al.~\cite{peng2019} who constrain discriminator networks, we can optimize (\ref{eq:info_elbo}) using dual gradient decent. Overall, we arrive at our final objective $L(p_\theta,q_\phi)$ by inserting the relaxation constraint into (\ref{eq:constr_elbo}) and introducing a scalar coefficient $\gamma_C$ and the Lagrange multiplier $\gamma_{\text{KL}}$ (which is still optimized via dual gradient decent), i.e.
\begin{equation}
\begin{aligned}
    L(p_\theta,q_\phi) = & \mathbb{E}_{q_\phi(z_\beta|\boldsymbol{x},x_t)} \left[ \log p_\theta(\boldsymbol{x}|z_\beta,x_t) \right] \\
    & - \gamma_{\text{KL}} \; (D_{\text{KL}}(q_\phi(z_\beta|\boldsymbol{x},x_t)||p(z_\beta)) - I_{\text{KL}}) \\
    & - \gamma_{C} \; \mathbb{E}_{q_\phi(z_\beta|\boldsymbol{x},x_t)} \left[\log \hat{p}_\psi(\boldsymbol{x}|z_\beta) \right] \; .
    \label{eq:final_elbo}
\end{aligned}
\end{equation}
\noindent
Note, that without our explicit disentanglement, relaxing $D_\text{KL}(q_\phi(z_\beta|\boldsymbol{x}, x_t)||p(z_\beta))$ would further encourage the entanglement of posture and behavior dynamics in $z_\beta$.\\
Relaxing the regularization $D_{\text{KL}}(q_\phi(z_\beta|\boldsymbol{x},x_t)||p(z_\beta))$ comes at the cost of a reduced overlap between the variational posterior $q_\phi$ and prior $p(z_\beta)$ impairing the sampling ability of our model. Next, we correct this missmatch by means of a subsequently learned normalizing flow transformation~\cite{glow,realnvp}.

\subsection{Bridging the gap between prior and posterior} 
We want to use our model not only to transfer behavior between videos, but also to synthesize novel behavior based on sampling $z_\beta$ from the prior distribution. Thus, strong deviations of the posterior $q_\phi(z_\beta|\boldsymbol{x},x_t)$ from $p(z_\beta)$ may reduce the syntheses results due to out-of-distribution samples. To alleviate this issue, we train a normalizing flow model~\cite{PapamakariosNormalizingFF,glow} after our variational behavior model is optimized. Normalizing flows yield an explicit, invertible transformation from $q_\phi$ to $p(z_\beta)$, thus bridging any potential gap between them. To this end, these models learn flexible probability distributions $p_u(u)$ over continuous random variables, such as our behavior representation $z_\beta$. In particular, normalizing flows establish a bijective mapping $z_\beta \overset{\mathcal{T}_\xi}{\longleftrightarrow} u$ using the transformation $\mathcal{T}_\xi = h_{\xi_1} \circ h_{\xi_2} \circ \dots \circ h_{\xi_m}$, a sequence of $m$ invertible functions $h_{\xi_j}$ parametrized by $\xi_j$ by maximizing the likelihood
\begin{equation}
    \mathbb{E}_{q_\phi(z_\beta|\boldsymbol{x},x_t)} \left[\log p_u(\mathcal{T}_\xi(z_\beta)) - \log |\det J_{\mathcal{T}_\xi}(z_\beta)|\right] \;.
\end{equation}
\noindent
Here, $\det J_{\mathcal{T}_\xi}$ is the Jacobian determinant of the invertible transformation. Choosing $p_u(u)$ to follow the same distribution as $p(z_\beta)$ establishes our desired bijective mapping between $q_\phi(z_\beta|\boldsymbol{x},x_t)$ and $p(z_\beta)$. Sampling novel behavior representations $z_\beta$ is then performed by $z_\beta = \mathcal{T}^{-1}_\xi(u), u \sim p_u(u)$. 

\begin{table*}
    \resizebox{\textwidth}{!}
    {
    \begin{tabular}{l||cc|cc|cc|cc|cc|cc||c|c}
    \toprule
     \multirow{2}{*}{Method} &
     \multicolumn{2}{c}{T$=$1} &
     \multicolumn{2}{c}{T$=$10} &
     \multicolumn{2}{c}{T$=$20} &
     \multicolumn{2}{c}{T$=$30} &
     \multicolumn{2}{c}{T$=$40} &
     \multicolumn{2}{c}{T$=$50} &
      \multicolumn{1}{c}{$acc.$} &
      \multicolumn{1}{c}{$d_{\beta}$} \\ 
       & {RE} & {TDE} & {RE} & {TDE}& {RE} & {TDE}& {RE} & {TDE}& {RE} & {TDE}& {RE} & {TDE} & {gt: 0.45} & {$\mu\pm \sigma$}\\
      \midrule
    cAE &  0.72 & 8.30 & 0.28 & 1.94 & 0.26 & 0.34 & 0.28 & 0.23 & 0.30 & 0.23 & 0.33 & 0.23 & 0.45 & 0.92$\pm$0.34 \\
    cVAE & 5.29 & 9.07 & 5.28 & 8.95 & 5.05 & 8.81 & 4.82 & 8.87 & 4.55 & 8.86 & 4.46 & 8.80  & 0.13 & 0.00 $\pm$ 0.00 \\
    MT-VAE~\cite{MT-VAE_YAN} & 1.36 & 8.90&1.40& 8.95& 1.39& 8.66& 1.38 &8.45& 1.34& 8.27& 1.37& 8.12& 0.20 & 4.44 $\pm$ 2.05 \\
    Ours ($\gamma_{C} = 0,I_\text{KL}=50$) & 1.71 & 9.01 & 1.46 & 7.92 & 1.22 & 6.95 & 1.17 & 6.15 & 1.18 & 5.58 & 1.30 & 5.33  & 0.35 & 2.82 $\pm$ 0.79 \\
    Ours ($\gamma_{C} = 0,I_\text{KL}=100$) & 1.24 & 8.99 & 0.89 & 7.09 & 0.81 & 5.55 & 0.78& 4.33 & 0.73 & 3.48 & 0.80 & 3.13 & 0.39 & 3.47 $\pm$ 0.93 \\
    Ours ($\gamma_{C} = 0,I_\text{KL}=200$) & 1.01 & 8.92 & 0.67 & 5.93 & 0.61 & 3.74 & 0.59 & 2.29 & 0.58 & 1.48 & 0.60 & 1.30 & 0.40 & 4.06 $\pm$1.18\\
    Ours ($I_\text{KL}=50$)  & 1.96 & 9.06 & 1.83 & 8.74 &1.74& 8.54 & 1.67 & 8.33 & 1.53 & 8.12 & 1.59 & 7.94 & 0.38 & 1.55 $\pm$ 0.61\\
    Ours ($I_\text{KL}=100$) & 2.01 & 9.08 & 1.96 & 8.78 & 1.88 & 8.57 & 1.76 & 8.37 & 1.77 & 8.15 & 1.76 & 8.0 & 0.38 & 1.60 $\pm$0.78\\
    Ours ($I_\text{KL}=200$) & 1.62 &  9.06& 1.47& 8.97& 1.56& 8.90& 1.47&8.77 &1.47 &8.58 &1.38&8.36  & 0.39 & 1.58 $\pm$0.71\\
    \bottomrule
  \end{tabular}
  }
  \caption{\textit{Evaluation of Behavior Transfer}. We compare different models on the task of behavior transfer using different metrics. The regression error 'RE' denotes the mean squared error (MSE) when predicting the source behavior sequence $x_\beta$ from the learned behavior representation $z_\beta$ using a regression network trained on this task. 'TDE' refers to the total displacement error measured as the MSE between $x_\beta$ and the re-enactment $x_R$. 'acc' denotes action classifier accuracy when using the respective behavior representations $z_\beta$ as input. 'gt:0.45' denotes the accuracy of an action classifier directly trained in ground-truth keypoint sequences, thus representing an upper bound on performance. For the latent space distance $d_\beta$ between the encodings of the source behavior $x_\beta$ and the re-enactment $x_R$ we report mean and standard deviation. Each metric is evaluated at timesteps $T \in \{1,10,20,30,40,50 \}$. Since we have no ground-truth data available for behavior transfer we cannot directly measure transfer performance. Instead, in Sec. 4.2 we show how the interplay of these metrics allow to evaluate transfer performance.}
  \label{tab:transfer_analysis}
\end{table*}

\section{Experiments}
We now investigate the capabilities of the prop osed method to disentangle pose of a sequence from the underlying behavior. The resulting model is evaluated for the tasks of behavior transfer to different start poses and diverse sampling from the behavior representation. Evaluation is performed on the \textit{Human3.6m} dataset \cite{h36m_pami}, a large-scale motion capture dataset which contains 3.6 million video frames of 11 subjects, each of which performs 17 actions. Following previous work \cite{kitani_dpp,yuan2020dlow} we use a 17-joint skeleton of 3D joint locations for training on 5 (S1,S5,S6,S7,S8) and testing on two subjects (S9,S11). We refer the reader to the supplementary or project page\footnote{\label{project_page}\url{https://cutt.ly/5l7rXEp}} for video material.

\subsection{Architecture and implementation details}
For the task of human behavior transfer, we use sequences of 50 frames as input for our network. The encoder-decoder networks representing $q_\phi(z_\beta|\boldsymbol{x},x_t)$ and $p_\theta(\boldsymbol{x}|z_\beta,x_t)$ are both implemented as a single-layer LSTM \cite{lstm} with a hidden dimensionality of 1024. Mean and variance of $q_\phi(z_\beta | \boldsymbol{x}, x_t)$ are realized as linear layers based on the final hidden state of the encoder. For our decoder $p_\theta(\boldsymbol{x}|z_\beta,x_t)$ we initialize the hidden state with the behavior representation $z_\beta$. The target posture $x_t$ is the input state of the decoder at the first time step. Subsequently the decoder uses its own predictions from the previous time step as input. For generating the individual postures, we follow \cite{martinez} and use a single linear layer on top of the LSTM output combined with residual skip connection to the input. The generative model $\hat{p}_\psi$ is implemented as a three-layer MLP to predict postures $\boldsymbol{x}$ given $z_\beta$. We model $p_\theta(\boldsymbol{x}|z_\beta, x_t)$ and $p_\psi(\boldsymbol{x}|z_\beta)$ as Gaussian, thus the expectations in Eq. \ref{eq:final_elbo} translate to mean squared errors. We train the network for 50 epochs and set $\gamma_C = 0.1$ and $I_{\text{KL}} = 100$ as discussed in the quantitative evaluation. 

\noindent \paragraph{Normalizing flow model $\mathcal{T}_\xi$.}
Our normalizing flow model $\mathcal{T}_\xi$ is implemented as a stacked sequence of 15 invertible neural networks based on an input dimensionality of $D=1024$. Each consists of 3 blocks of subsequently applied actnorm \cite{glow}, affine coupling layers \cite{coupling_lay} and shuffling layers. The affine coupling layers consist of 2 fully connected layers with dimensionality $D=1024$. We trained the normalizing flow model on a single Titan Xp for 5 epochs with batchsize $64$ and ADAM~\cite{adam} optimizer with learning rate $6.5 \times 10^{-6}$. Further information regarding our normalizing flow model is provided in the Appendix \ref{sec:impl_details}.

\noindent \paragraph{Model for posture-appearance transfer.}
In order to be able to synthesize realistic RGB videos of human behavior, we translate our generated postures to RGB images. To this end, we utilize our proposed framework for the task of shape and appearance disentanglement \cite{app_shape_dis_2}. We train a model to obtain an appearance representation from static images, which is independent of the corresponding posture information. Thus, we can use our method to transfer behavior from a source sequence to a given target posture and generate an animated video sequence by frame-wise synthesizing RGB images. More details on our posture and appearance model and further results can be found in the Appendix \ref{sec:shape_appearance}. 

\subsection{Behavior re-enactment}
We now evaluate our proposed model qualitatively and quantitatively for the task of behavior transfer and its abilities to sample and synthesize novel behavior. 

\noindent \paragraph{Qualitative evaluation.}
Figure \ref{fig:behavior_transfer} shows examples of transferred behavior. We show the posture sequence $x_\beta$ exhibiting a source behavior $\beta$ (top row) and its transfer to different, unrelated postures $x_t$. The re-enactments depict both the re-enacted posture sequence and the rendered RGB video frames based on the model for posture and appearance transfer. Since our model captures behavior independent of posture, it successfully transfers only the characteristic body dynamics of $\beta$ and infers potentially needed transitions itself. As a result, the target posture $x_t$ is naturally animated to perform behavior $\beta$ independent of diverse target postures, such as standing or sitting.
For instance, in the example at the left top, each person accurately raises both hands to its head. Note that, in the last example on the top left, the person does not change its posture since the hands are already up. Moreover, the kneeling person on the bottom left only lowers its torso as its knees are already placed on the ground and cannot be bent further. More visual examples can be found in the Appendix \ref{sec:add_results}. Video examples are provided on our project page\textsuperscript{\ref{project_page}}.

\begin{table}[t]
\centering
    \begin{tabular}[t]{l|cccc}
    \toprule
    \multirow{1}{*}{Method} & 
    \multicolumn{4}{c}{Type of synthesis} \\
    & {self} & {transfer} & {prior} & {flow}  \\
        \midrule
        MT-VAE~\cite{MT-VAE_YAN} & \textbf{0.49} & 0.17 & -& -\\
         Ours ($\gamma_{C} = 0$) & \textbf{0.49} & 0.14 & 0.09 & 0.12 \\
         Ours & \textbf{0.49} & \textbf{0.23} & \textbf{0.13} & \textbf{0.23} \\
         \bottomrule
    \end{tabular}
\caption{\textit{Realism of behavior generations.} Evaluation of sampling quality using a discriminator classifying between ground-truth sequences and behavior generations of different origins: 'self' denotes video reconstructions, 'transfer' denotes generations depicting actual behavior transfer, i.e. using a randomly sampled starting posture and a extracted behavior $q_\phi(z_\beta|\boldsymbol{x},x_t)$, 'prior' denotes synthesizing behavior sampled from the prior representation $p(z_\beta)$ and 'flow' refers to synthesizing behavior sampled using the invertible mapping $\mathcal{T}_\xi$.}
\label{fig:sampling_quality}
\vspace{-5mm}
\end{table}

\noindent \paragraph{Quantitative evaluation.} We now evaluate how well our model transfers behavior $\beta$ extracted from a source sequence $x_\beta$ to an initial, unrelated target posture $x_t$ taken from random, different sequences. Meaningful re-enactment of $\beta$ should only transfer characteristic body dynamics to a target posture $x_t$. We compare our model to different baseline models, i.e. conditional autoencoder (cAE), vanilla conditional variational autoencoder (cVAE) and our model with and without our proposed posture disentanglement. Each model uses the same architectures except for deviations due to individual training objectives: The cAE is trained without disentanglement and without variational bottleneck. The cVAE is trained on the training objective Eq. (\ref{eq:elbo}). Moreover, we compare the MT-VAE~\cite{MT-VAE_YAN} which uses latent space arithemtics to transform between different actions. \\
A model failing this task would typically generate \textit{(i)} sequences which rather exactly copy full postures of the source sequence $\boldsymbol{x}_\beta$ in contrast to transferring only its characteristic dynamics to $x_t$; or \textit{(ii)} generating behavior different to $\beta$ such as some likely future behavior of $x_t$. To identify \textit{(i)}, we measure the transfer displacement error (TDE), i.e. the displacement error between postures of the re-enactment $\boldsymbol{x}_R$ and source $\boldsymbol{x}_\beta$ at time-steps $T$. For \textit{(ii)}, since we have no ground-truth available for behavior transfer, we measure the average euclidean distance $d_\beta$ in the representation space $z_\beta$ between encodings of $\boldsymbol{x}_\beta$ and $\boldsymbol{x}_R$. Combined, a well transferring model should yield re-enacted sequences $\boldsymbol{x}_R$ which is dissimilar to $\boldsymbol{x}_\beta$, thus not merely copying postures (i.e. large TDE). Given this, both sequences should be similar in representation space (i.e. small $d_\beta$) to indicate their similarity in behavior. Moreover, the representations need to be informative to exclude degenerated solutions. For the latter we examine their benefit for action classification on H3.6M (acc.). For each experiment we provide detailed protocols and implementation details in the Appendix \ref{sec:abl_protocoll}. \\
Tab.~\ref{tab:transfer_analysis} evaluates these experiments. We observe that cAE exhibits a strong decline in TDE values as $T$ increases, resulting in TDE values close to $0$. Thus, this model accurately copies the posture sequence $x_\beta$, instead of inferring behavior $\beta$ and potentially needed transitions itself (cf. suppl. video material). Consequently, its behavior representation only captures posture information, rather than body dynamics. In contrast, the cVAE model consistently reaches high TDE scores, thus generating posture sequences which are very different from $x_\beta$. However, the distance $d_\beta$ shows that the model suffers from posterior collapse, hence $z_\beta$ is neglected and only likely continuations following $x_t$ are predicted (see also Appendix \ref{sec:abl_protocoll}). Relaxing the information bottleneck of cVAE (i.e. our model without disentanglement, Eq.(\ref{eq:info_elbo})) alleviates the posterior collapse and $z_\beta$ becomes informative. Looking at different settings for $I_{\text{KL}}$, we see that TDE values slowly decrease with $T$ and ranges between cAE and cVAE, while exhibiting large values of $d_\beta$. We attribute this to a distorted latent representation being learned. In contrast, our full model with explicit disentanglement of behavior and posture exhibits large TDE values matching those of the cVAE while at the same time mapping $\boldsymbol{x}_\beta$ and $\boldsymbol{x}_R$ close in $z_\beta$. Thus, since postures are very different, closeness in $z_\beta$ arises from similarity in body dynamics, highlighting actual behavior transfer. The accuracy of the action classifier (\textit{acc.}) confirms that the captured dynamics are informative, almost matching the classifier result when training on ground truth sequences. Moreover, the classifier performance for MT-VAE reveals that their latent space is significantly less informative than our behavior representation, which indicates limited motion transfer capabilities. Indeed, a large mean distance $d_\beta$ of $4.44$ shows a strong difference in representation between source and re-enacted behavior, most likely due to the linear arithemitcs assumption not being hold. \\
\begin{table}[t]
\centering
    \begin{tabular}{l|cc|cc}
    \toprule
    \multirow{2}{*}{Method} & 
    \multicolumn{2}{c}{N$=$10} &
    \multicolumn{2}{c}{N$=$50} \\
    & {ASD} & {FSD} &{ASD} & {FSD}   \\
    \midrule
         cVAE \cite{kitani_dpp} & 0.25& 0.36 & 0.16 & 0.22\\
         DSF \cite{kitani_dpp} & 0.38 & 0.62 & 0.31 & 0.42\\
         Ours & \textbf{0.63} & \textbf{0.88} & \textbf{0.45} & \textbf{0.58}\\
     \bottomrule
    \end{tabular}
\caption{\textit{Evaluation of Sampling Capabilities.} (a) Quantitative evaluation of diversity with ASD and FSD \cite{kitani_dpp}, numbers are taken from \cite{kitani_dpp}.}
\label{fig:3lettermetrics}
\vspace{-3mm}
\end{table}
To provide an additional, explicit measure for disentanglement of behavior and posture, we adapt an evaluation procedure inspired by works on identifying latent factors of variation~\cite{mathieu16}. To this end, we train a regression network to predict posture coordinates of $x_\beta$ from its encoding $z_\beta$ at different time-steps and report the average regression errors (RE) in Tab.~\ref{tab:transfer_analysis}. Naturally, the cAE model results in very low errors due to copying, while the cVAE exhibits large REs due to the posterior collapse. Comparing our model with and without disentanglement demonstrates consistently higher prediction errors, indicating that only few posture information is encoded in $z_\beta$. Moreover, our analysis shows that our model is robust to the choice of $I_{\text{KL}}$. In the remainder of the experiments we choose $I_{\text{KL}}=100$.


\begin{figure}[b]
    \centering
    \includegraphics[width=0.99\columnwidth]{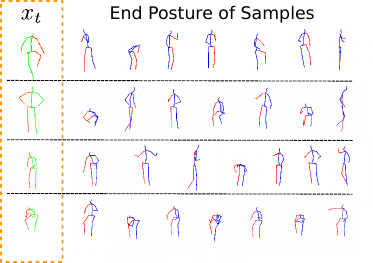}
    \caption{Qualitative visualization of diversity by showing the end poses from our sampled behaviors.}
    \label{fig:my_label}
\end{figure}
\subsection{Sampling and synthesis of novel behavior} 
We now evaluate our model on the task of synthesizing novel behavior by sampling behavior representations $z_\beta$ from the prior distribution $p(z_\beta)$. Following other approaches for human motion synthesis \cite{yuan2020dlow, kitani_dpp, mix-and-match} we evaluate the aspect of sampling quality~\cite{mix-and-match} and diversity~\cite{yuan2020dlow,kitani_dpp}. To address the first we train a binary classifier to distinguish between 25k ground-truth and 25k generated sequences. The accuracy of the classifier determines the realism of the evaluated samples and is reported in Tab.~\ref{fig:sampling_quality}. The implementation details for the classifier can be found in the Appendix \ref{sec:abl_protocoll}. Posture sequences synthesized using the explicit invertible mapping $\mathcal{T}_\xi$ between prior and posterior $q_\phi(z_\beta|\boldsymbol{x},x_t)$ are more realistic than directly using prior samples. This is explained by the corrected mismatch between posterior and prior distribution and clearly demonstrated by the visual comparisons in the videos contained in the accompanying video material. Moreover, we observe that our explicit disentanglement of posture and behavior significant improves the quality of samples. In particular, we also outperform MT-VAE~\cite{MT-VAE_YAN} by relative $35\%$ in behavior re-enactment ('transfer'). \\
For evaluating diversity we follow the evaluation protocol of \cite{yuan2020dlow, kitani_dpp} by using the following metrics:
\textit{(i)} Average Self Distance (ASD): Average euclidean distance between a generated sequence and its closest neighbor sequence among generations; and \textit{(ii)} Final Self Distance (FSD): Euclidean distance between the last posture of a generated sequence and its closest neighbor’s final posture. 
Both ASD and FSD assess the uniqueness of samples. Tab.~\ref{fig:3lettermetrics} compares ASD and FSD scores of our model with the cVAE and the diversity sampler function (DSF) from \cite{kitani_dpp} for sample-set sizes of $K \in \{10,50\}$.
For both metrics, we outperform the other approaches by a significant margin.
Finally, we visually demonstrate the diversity of our samples in Fig~\ref{fig:my_label} by showing the final postures of sampled behaviors.

\section{Discussion}
We presented a conditional generative model for controlled synthesis and transfer of human behavior. To this end, we learn a dedicated representation for human behavior disentangled from posture. By extracting the characteristic body dynamics from a video depicting a certain behavior, our model is able to animate persons observed in significantly different postures. A particular challenge arises from animating postures which allow for no direct transfer of behavior dynamics, but require an intermediate transition. Correct inference of such transition is essentially a generalization problem asking for synthesis outside the training distribution. While our model successfully infers such transitions to a certain degree, it fails in cases of complex transitions needed, such as enacting a walking behavior by a person which is lying on the ground. This shows that our introduced problem requires a deep understanding of behavior, thus posing a new challenge for research on human motion synthesis in general.

\section*{Acknowledgements}
The research leading to these results is funded by the German Federal Ministry for Economic Affairs and Energy within the project “KI-Absicherung – Safe AI for automated driving” and by the German Research Foundation (DFG) within project 421703927.

{\small
\bibliographystyle{ieee_fullname}
\bibliography{bibliography}

\begin{thebibliography}{10}\itemsep=-1pt

\bibitem{mix-and-match}
Mohammad~Sadegh Aliakbarian, Fatemeh~Sadat Saleh, Mathieu Salzmann, Lars
  Petersson, Stephen Gould, and Amirhossein Habibian.
\newblock Learning variations in human motion via mix-and-match perturbation,
  2019.

\bibitem{recycleGAN}
Aayush Bansal, Shugao Ma, Deva Ramanan, and Yaser Sheikh.
\newblock Recycle-gan: Unsupervised video retargeting.
\newblock In {\em ECCV}, 2018.

\bibitem{HPgan}
Emad Barsoum, John Kender, and Zicheng Liu.
\newblock Hp-gan: Probabilistic 3d human motion prediction via gan.
\newblock {\em IEEE Conference on Computer Vision and Pattern Recognition
  Workshops (CVPRW)}, 2017.

\bibitem{mine}
Mohamed~Ishmael Belghazi, Aristide Baratin, Sai Rajeshwar, Sherjil Ozair,
  Yoshua Bengio, Aaron Courville, and Devon Hjelm.
\newblock Mutual information neural estimation.
\newblock In {\em Proceedings of the 35th International Conference on Machine
  Learning}, 2018.

\bibitem{bowman16}
Samuel~R. Bowman, Luke Vilnis, Oriol Vinyals, Andrew Dai, Rafal Jozefowicz, and
  Samy Bengio.
\newblock Generating sentences from a continuous space.
\newblock In {\em Proceedings of The 20th {SIGNLL} Conference on Computational
  Natural Language Learning}, Aug. 2016.

\bibitem{uBAM_Biagio}
Biagio Brattoli, Uta B{\"{u}}chler, Michael Dorkenwald, Philipp Reiser, Linard
  Filli, Fritjof Helmchen, Anna{-}Sophia Wahl, and Bj{\"{o}}rn Ommer.
\newblock ubam: Unsupervised behavior analysis and magnification using deep
  learning.
\newblock {\em CoRR}, abs/2012.09237, 2020.

\bibitem{biggan}
Andrew Brock, Jeff Donahue, and Karen Simonyan.
\newblock Large scale {GAN} training for high fidelity natural image synthesis.
\newblock In {\em International Conference on Learning Representations (ICLR)},
  2019.

\bibitem{carreira2018quo}
Joao Carreira and Andrew Zisserman.
\newblock Quo vadis, action recognition? a new model and the kinetics dataset,
  2018.

\bibitem{dance_vid2vid}
Caroline Chan, Shiry Ginosar, Tinghui Zhou, and Alexei~A Efros.
\newblock Everybody dance now.
\newblock In {\em IEEE International Conference on Computer Vision (ICCV)},
  2019.

\bibitem{lossyVAE}
Xi Chen, P.~Diederik Kingma, Tim Salimans, Yan Duan, Prafulla Dhariwal, John
  Schulman, Ilya Sutskever, and Pieter Abbeel.
\newblock Variational lossy autoencoder.
\newblock {\em ICLR}, 2017.

\bibitem{ActionAgnosticHP}
Hsu-Kuang Chiu, Ehsan Adeli, Borui Wang, De-An Huang, and Juan~Carlos Niebles.
\newblock Action-agnostic human pose forecasting.
\newblock {\em 2019 IEEE Winter Conference on Applications of Computer Vision
  (WACV)}, 2018.

\bibitem{datta2002person}
Ankur Datta, Mubarak Shah, and N~Da~Vitoria Lobo.
\newblock Person-on-person violence detection in video data.
\newblock In {\em Object recognition supported by user interaction for service
  robots}. IEEE, 2002.

\bibitem{app_shape_dis_2}
Rodrigo de Bem, Arnab Ghosh, Thalaiyasingam Ajanthan, Ondrej Miksik, N.
  Siddharth, and Philip H.~S. Torr.
\newblock Dgpose: Disentangled semi-supervised deep generative models for human
  body analysis.
\newblock In {\em abs/1804.06364}, 2018.

\bibitem{unsup_ffp}
Emily~L Denton and vighnesh Birodkar.
\newblock Unsupervised learning of disentangled representations from video.
\newblock In {\em Advances in Neural Information Processing Systems}, 2017.

\bibitem{realnvp}
Laurent Dinh, Jascha Sohl-Dickstein, and Samy Bengio.
\newblock Density estimation using real nvp.
\newblock In {\em international conference on learning representations}, 2017.

\bibitem{coupling_lay}
Laurent Dinh, Jascha Sohl-Dickstein, and Samy Bengio.
\newblock Density estimation using real nvp.
\newblock 2017.

\bibitem{magnification_dorkenwald}
Michael Dorkenwald, Uta B{\"{u}}chler, and Bj{\"{o}}rn Ommer.
\newblock Unsupervised magnification of posture deviations across subjects.
\newblock In {\em 2020 {IEEE/CVF} Conference on Computer Vision and Pattern
  Recognition, {CVPR} 2020, Seattle, WA, USA, June 13-19, 2020}, pages
  8253--8263. {IEEE}, 2020.

\bibitem{hbugen}
Patrick Esser, Johannes Haux, Timo Milbich, and Bj{\"o}rn Ommer.
\newblock Towards learning a realistic rendering of human behavior.
\newblock In {\em European Conference on Computer Vision - Workshops}, 2018.

\bibitem{esser2019unsupervised}
Patrick Esser, Johannes Haux, and Bj{\"o}rn Ommer.
\newblock Unsupervised robust disentangling of latent characteristics for image
  synthesis.
\newblock In {\em Proceedings of the Intl. Conf. on Computer Vision (ICCV)},
  2019.

\bibitem{vunet}
Patrick Esser, Ekaterina Sutter, and Bj{\"o}rn Ommer.
\newblock A variational u-net for conditional appearance and shape generation.
\newblock In {\em IEEE Conference on Computer Vision and Pattern Recognition},
  2018.

\bibitem{Fragkiadaki2015RecurrentNM}
Katerina Fragkiadaki, Sergey Levine, Panna Felsen, and Jitendra Malik.
\newblock Recurrent network models for human dynamics.
\newblock {\em International Conference on Computer Vision (ICCV)}, 2015.

\bibitem{game_vid2vid}
Oran Gafni, Lior Wolf, and Yaniv Taigman.
\newblock Vid2game: Controllable characters extracted from real-world videos.
\newblock In {\em International Conference on Learning Representations (ICLR)},
  2020.

\bibitem{gebert2019end}
Patrick Gebert, Alina Roitberg, Monica Haurilet, and Rainer Stiefelhagen.
\newblock End-to-end prediction of driver intention using 3d convolutional
  neural networks.
\newblock In {\em 2019 IEEE Intelligent Vehicles Symposium (IV)}. IEEE, 2019.

\bibitem{gan}
Ian Goodfellow, Jean Pouget-Abadie, Mehdi Mirza, Bing Xu, David Warde-Farley,
  Sherjil Ozair, Aaron Courville, and Yoshua Bengio.
\newblock Generative adversarial nets.
\newblock In {\em Advances in Neural Information Processing Systems 27}, 2014.

\bibitem{controllableVidGen}
Zekun Hao, Xun Huang, and Serge Belongie.
\newblock Controllable video generation with sparse trajectories.
\newblock In {\em IEEE Conference on Computer Vision and Pattern Recognition
  (CVPR)}, 2018.

\bibitem{cycleVAE}
Ananya Harsh~Jha, Saket Anand, Maneesh Singh, and V.~S.~R. Veeravasarapu.
\newblock Disentangling factors of variation with cycle-consistent variational
  auto-encoders.
\newblock In {\em Proceedings IEEE European Conference on Computer Vision},
  2018.

\bibitem{lstm}
Sepp Hochreiter and J{\"u}rgen Schmidhuber.
\newblock Long short-term memory.
\newblock {\em Neural computation}, 9(8):1735--1780, 1997.

\bibitem{h36m_pami}
Catalin Ionescu, Dragos Papava, Vlad Olaru, and Cristian Sminchisescu.
\newblock Human3.6m: Large scale datasets and predictive methods for 3d human
  sensing in natural environments.
\newblock {\em IEEE Transactions on Pattern Analysis and Machine Intelligence},
  2014.

\bibitem{struct_RNN}
A. {Jain}, A.~R. {Zamir}, S. {Savarese}, and A. {Saxena}.
\newblock Structural-rnn: Deep learning on spatio-temporal graphs.
\newblock In {\em 2016 IEEE Conference on Computer Vision and Pattern
  Recognition (CVPR)}, 2016.

\bibitem{prog_grow}
Tero Karras, Timo Aila, Samuli Laine, and Jaakko Lehtinen.
\newblock Progressive growing of gans for improved quality, stability, and
  variation.
\newblock In {\em International Conference on Learning Representations (ICLR)},
  2018.

\bibitem{stylegan}
Tero Karras, Samuli Laine, and Timo Aila.
\newblock A style-based generator architecture for generative adversarial
  networks.
\newblock In {\em IEEE Conference on Computer Vision and Pattern Recognition},
  2019.

\bibitem{adam}
Diederik~P. Kingma and Jimmy Ba.
\newblock Adam: {A} method for stochastic optimization.
\newblock In Yoshua Bengio and Yann LeCun, editors, {\em 3rd International
  Conference on Learning Representations, {ICLR} 2015, San Diego, CA, USA, May
  7-9, 2015, Conference Track Proceedings}, 2015.

\bibitem{glow}
Durk~P Kingma and Prafulla Dhariwal.
\newblock Glow: Generative flow with invertible 1x1 convolutions.
\newblock In {\em Advances in Neural Information Processing Systems 31}, 2018.

\bibitem{vae}
Diederik~P Kingma and Max Welling.
\newblock Auto-encoding variational bayes, 2014.

\bibitem{aegan}
Anders Boesen~Lindbo Larsen, S\o{}ren~Kaae S\o{}nderby, Hugo Larochelle, and
  Ole Winther.
\newblock Autoencoding beyond pixels using a learned similarity metric.
\newblock In {\em Proceedings of the 33rd International Conference on
  International Conference on Machine Learning - Volume 48}, 2016.

\bibitem{lwb2019}
Wen Liu, Zhixin Piao, Min Jie, Wenhan Luo, Lin Ma, and Shenghua Gao.
\newblock Liquid warping gan: A unified framework for human motion imitation,
  appearance transfer and novel view synthesis.
\newblock In {\em The IEEE International Conference on Computer Vision (ICCV)},
  2019.

\bibitem{deepfashion_data}
Ziwei Liu, Ping Luo, Shi Qiu, Xiaogang Wang, and Xiaoou Tang.
\newblock Deepfashion: Powering robust clothes recognition and retrieval with
  rich annotations.
\newblock In {\em The IEEE Conference on Computer Vision and Pattern
  Recognition (CVPR)}, June 2016.

\bibitem{unsup_partbased}
Dominik Lorenz, Leonard Bereska, Timo Milbich, and Bj{\"o}rn Ommer.
\newblock Unsupervised part-based disentangling of object shape and appearance.
\newblock In {\em Proceedings of the IEEE Conference on Computer Vision and
  Pattern Recognition (CVPR)}, 2019.

\bibitem{app_shape_dis_1}
Liqian Ma, Xu Jia, Qianru Sun, Bernt Schiele, Tinne Tuytelaars, and Luc
  Van~Gool.
\newblock Pose guided person image generation.
\newblock In {\em Advances in Neural Information Processing Systems}, 2017.

\bibitem{ma2017disentangled}
Liqian Ma, Qianru Sun, Stamatios Georgoulis, Luc Van~Gool, Bernt Schiele, and
  Mario Fritz.
\newblock Disentangled person image generation.
\newblock In {\em The IEEE International Conference on Computer Vision and
  Pattern Recognition (CVPR)}, June 2018.

\bibitem{martin2019drive}
Manuel Martin, Alina Roitberg, Monica Haurilet, Matthias Horne, Simon Rei{\ss},
  Michael Voit, and Rainer Stiefelhagen.
\newblock Drive\&act: A multi-modal dataset for fine-grained driver behavior
  recognition in autonomous vehicles.
\newblock In {\em Proceedings of the IEEE International Conference on Computer
  Vision}, 2019.

\bibitem{martinez}
Julieta Martinez, Michael~J. Black, and Javier Romero.
\newblock On human motion prediction using recurrent neural networks.
\newblock In {\em Proceedings IEEE Conference on Computer Vision and Pattern
  Recognition (CVPR) 2017}, Piscataway, NJ, USA, July 2017. IEEE.

\bibitem{masood2018detecting}
Sarfaraz Masood, Abhinav Rai, Aakash Aggarwal, Mohammad~Najmud Doja, and
  Musheer Ahmad.
\newblock Detecting distraction of drivers using convolutional neural network.
\newblock {\em Pattern Recognition Letters}, 2018.

\bibitem{mathieu16}
Michael~F Mathieu, Junbo~Jake Zhao, Junbo Zhao, Aditya Ramesh, Pablo
  Sprechmann, and Yann LeCun.
\newblock Disentangling factors of variation in deep representation using
  adversarial training.
\newblock In {\em Advances in Neural Information Processing Systems}, 2016.

\bibitem{umap}
Leland McInnes and John Healy.
\newblock Umap: Uniform manifold approximation and projection for dimension
  reduction.
\newblock 2018.

\bibitem{iccv17_unsup_video}
Timo Milbich, Miguel Bautista, Ekaterina Sutter, and Bj{\"o}rn Ommer.
\newblock Unsupervised video understanding by reconciliation of posture
  similarities.
\newblock In {\em Proceedings of the IEEE International Conference on Computer
  Vision}, 2017.

\bibitem{PapamakariosNormalizingFF}
George Papamakarios, Eric~T. Nalisnick, Danilo~Jimenez Rezende, Shakir Mohamed,
  and Balaji Lakshminarayanan.
\newblock Normalizing flows for probabilistic modeling and inference, 2019.

\bibitem{peng2019}
Bin~Xue Peng, Angjoo Kanazawa, Samuel Toyer, Pieter Abbeel, and Sergey Levine.
\newblock Variational discriminator bottleneck: Improving imitation learning,
  inverse rl, and gans by constraining information flow.
\newblock {\em international conference on learning representations}, 2019.

\bibitem{mutualAE}
Mary Phuong, Max Welling, Nate Kushman, Ryota Tomioka, and Sebastian Nowozin.
\newblock The mutual autoencoder: Controlling information in latent code
  representations, 2018.

\bibitem{rezaabad20}
Ali Rezaabad and Sriram Vishwanath.
\newblock Learning representations by maximizing mutual information in
  variational autoencoder, 2019.

\bibitem{densePose}
Iasonas~Kokkinos Riza Alp~Gueler, Natalia~Neverova.
\newblock Densepose: Dense human pose estimation in the wild.
\newblock In {\em The IEEE Conference on Computer Vision and Pattern
  Recognition (CVPR)}, 2018.

\bibitem{inception_score}
Tim Salimans, Ian Goodfellow, Wojciech Zaremba, Vicki Cheung, Alec Radford, Xi
  Chen, and Xi Chen.
\newblock Improved techniques for training gans.
\newblock In {\em Advances in Neural Information Processing Systems 29}. 2016.

\bibitem{vgg}
Karen Simonyan and Andrew Zisserman.
\newblock Very deep convolutional networks for large-scale image recognition.
\newblock In {\em International Conference on Learning Representations}, 2015.

\bibitem{cvae}
Kihyuk Sohn, Xinchen Yan, and Honglak Lee.
\newblock Learning structured output representation using deep conditional
  generative models.
\newblock In {\em International Conference on Neural Information Processing
  Systems}, 2015.

\bibitem{sultani2018real}
Waqas Sultani, Chen Chen, and Mubarak Shah.
\newblock Real-world anomaly detection in surveillance videos.
\newblock In {\em Proceedings of the IEEE Conference on Computer Vision and
  Pattern Recognition}, 2018.

\bibitem{tishby2015deep}
Naftali Tishby and Noga Zaslavsky.
\newblock Deep learning and the information bottleneck principle, 2015.

\bibitem{deep_feat_interpol}
P. {Upchurch}, J. {Gardner}, G. {Pleiss}, R. {Pless}, N. {Snavely}, K. {Bala},
  and K. {Weinberger}.
\newblock Deep feature interpolation for image content changes.
\newblock In {\em 2017 IEEE Conference on Computer Vision and Pattern
  Recognition (CVPR)}, 2017.

\bibitem{vidsynthVondrick}
Carl Vondrick, Hamed Pirsiavash, and Antonio Torralba.
\newblock Generating videos with scene dynamics.
\newblock In D.~D. Lee, M. Sugiyama, U.~V. Luxburg, I. Guyon, and R. Garnett,
  editors, {\em Advances in Neural Information Processing Systems 29}, 2016.

\bibitem{wahl2017optogenetically}
Anna-Sophia Wahl, U B{\"u}chler, A Br{\"a}ndli, Biagio Brattoli, Simon Musall,
  Hansj{\"o}rg Kasper, Benjamin~V Ineichen, Fritjof Helmchen, Bj{\"o}rn Ommer,
  and Martin~E Schwab.
\newblock Optogenetically stimulating intact rat corticospinal tract
  post-stroke restores motor control through regionalized functional circuit
  formation.
\newblock {\em Nature communications}, 2017.

\bibitem{theposeknows}
Jacob Walker, Kenneth Marino, Abhinav Gupta, and Martial Hebert.
\newblock The pose knows: Video forecasting by generating pose futures.
\newblock In {\em International Conference on Computer Vision}, 2017.

\bibitem{fewshot_vid2vid}
Ting-Chun Wang, Ming-Yu Liu, Andrew Tao, Guilin Liu, Jan Kautz, and Bryan
  Catanzaro.
\newblock Few-shot video-to-video synthesis.
\newblock In {\em Advances in Neural Information Processing Systems (NeurIPS)},
  2019.

\bibitem{vid2vid}
Ting-Chun Wang, Ming-Yu Liu, Jun-Yan Zhu, Guilin Liu, Andrew Tao, Jan Kautz,
  and Bryan Catanzaro.
\newblock Video-to-video synthesis.
\newblock In {\em Advances in Neural Information Processing Systems (NeurIPS)},
  2018.

\bibitem{ssim}
Zhou Wang, Alan~C. Bovik, Hamid~R. Sheikh, and Eero~P. Simoncelli.
\newblock Image quality assessment: From error visibility to structural
  similarity.
\newblock {\em IEEE TRANSACTIONS ON IMAGE PROCESSING}, 2004.

\bibitem{wu2018compressed}
Chao-Yuan Wu, Manzil Zaheer, Hexiang Hu, R. Manmatha, Alexander~J. Smola, and
  Philipp Krähenbühl.
\newblock Compressed video action recognition, 2018.

\bibitem{alphapose}
Yuliang Xiu, Jiefeng Li, Haoyu Wang, Yinghong Fang, and Cewu Lu.
\newblock {Pose Flow}: Efficient online pose tracking.
\newblock In {\em BMVC}, 2018.

\bibitem{xu2015learning}
Dan Xu, Elisa Ricci, Yan Yan, Jingkuan Song, and Nicu Sebe.
\newblock Learning deep representations of appearance and motion for anomalous
  event detection.
\newblock {\em arXiv preprint arXiv:1510.01553}, 2015.

\bibitem{MT-VAE_YAN}
Xinchen Yan, Akash Rastogi, Ruben Villegas, Kalyan Sunkavalli, Eli Shechtman,
  Sunil Hadap, Ersin Yumer, and Honglak Lee.
\newblock {MT-VAE:} learning motion transformations to generate multimodal
  human dynamics.
\newblock In Vittorio Ferrari, Martial Hebert, Cristian Sminchisescu, and Yair
  Weiss, editors, {\em Computer Vision - {ECCV} 2018 - 15th European
  Conference, Munich, Germany, September 8-14, 2018, Proceedings, Part {V}},
  volume 11209 of {\em Lecture Notes in Computer Science}, pages 276--293.
  Springer, 2018.

\bibitem{poseGuidedVidGen}
Ceyuan Yang, Zhe Wang, Xinge Zhu, Chen Huang, Jianping Shi, and Dahua Lin.
\newblock Pose guided human video generation.
\newblock In {\em ECCV}, 2018.

\bibitem{yuan2020dlow}
Ye Yuan and Kris Kitani.
\newblock Dlow: Diversifying latent flows for diverse human motion prediction,
  2020.

\bibitem{kitani_dpp}
Ye Yuan and Kris~M. Kitani.
\newblock Diverse trajectory forecasting with determinantal point processes.
\newblock In {\em 8th International Conference on Learning Representations,
  {ICLR} 2020}, 2020.

\bibitem{infoVAE}
Shengjia Zhao, Jiaming Song, and Stefano Ermon.
\newblock Infovae: Information maximizing variational autoencoders, 2017.

\bibitem{mark_data}
Liang Zheng, Liyue Shen, Lu Tian, Shengjin Wang, Jingdong Wang, and Qi Tian.
\newblock Scalable person re-identification: A benchmark.
\newblock In {\em Proceedings of the 2015 IEEE International Conference on
  Computer Vision (ICCV)}, ICCV ’15, page 1116–1124, USA, 2015. IEEE
  Computer Society.

\end{thebibliography}
}


\clearpage
\newpage

{\noindent \Huge \textbf{Appendix}}

\appendix

\section{Behavior Model}

\subsection{Training and Implementation Details}
\label{sec:impl_details}

\paragraph{Behavior model}
Most of the implementation details of our behavior model are already described in the main paper in Sec. 4. We train our model on a single Titan Xp using ADAM~\cite{adam} optimizer with learning rate $0.0001$ which is decreased after $10,25$ and $35$ epochs. For data preprocessing, we normalize the posture keypoints to have zero mean and unit variance.

\begin{figure*}[h]
\begin{minipage}[c]{.5\linewidth}
    \centering
    \includegraphics[width=\linewidth]{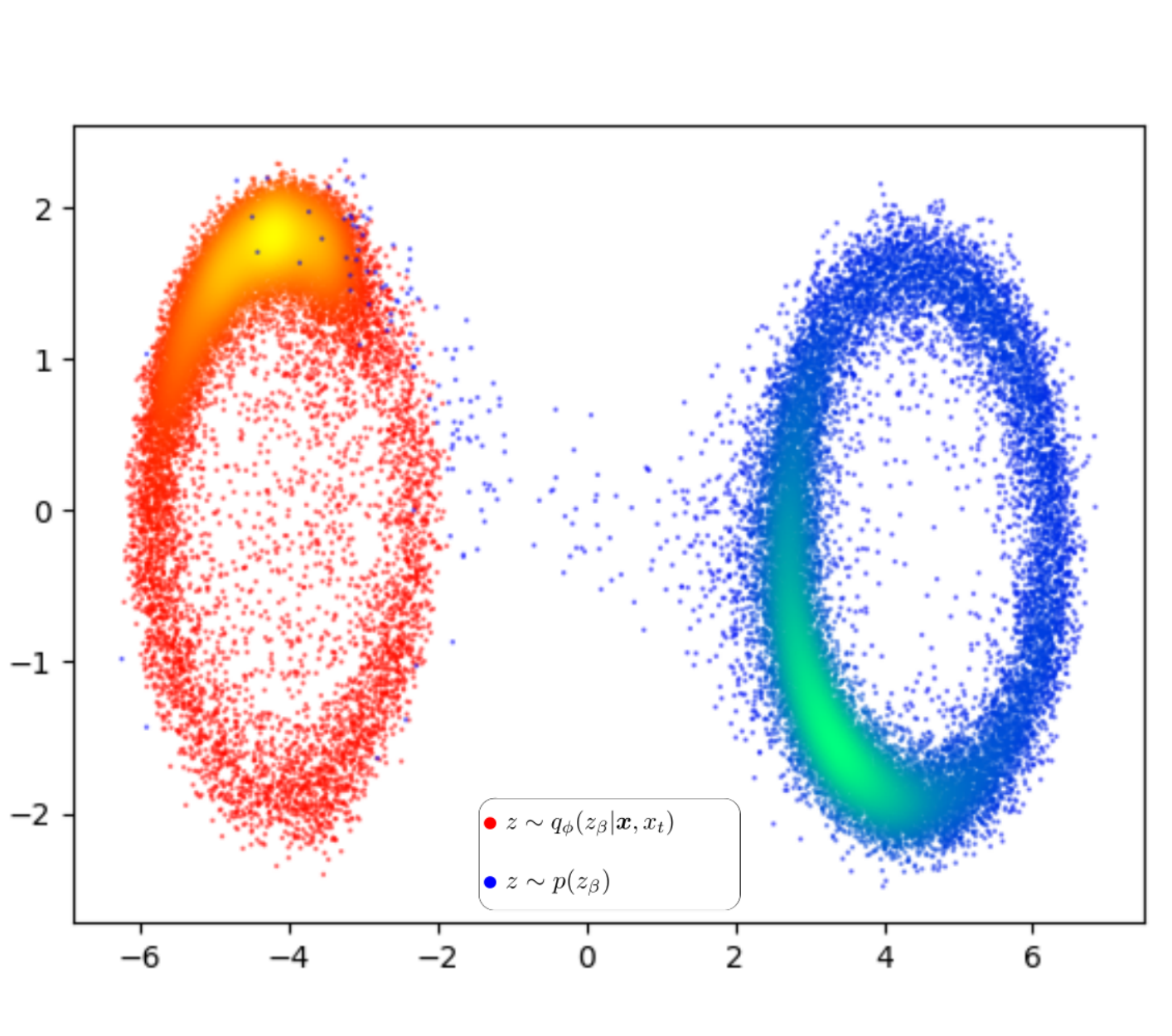}
    (a)
 \end{minipage}
 \hfill
 \begin{minipage}[c]{0.5\linewidth}
    \centering
    \includegraphics[width=\linewidth]{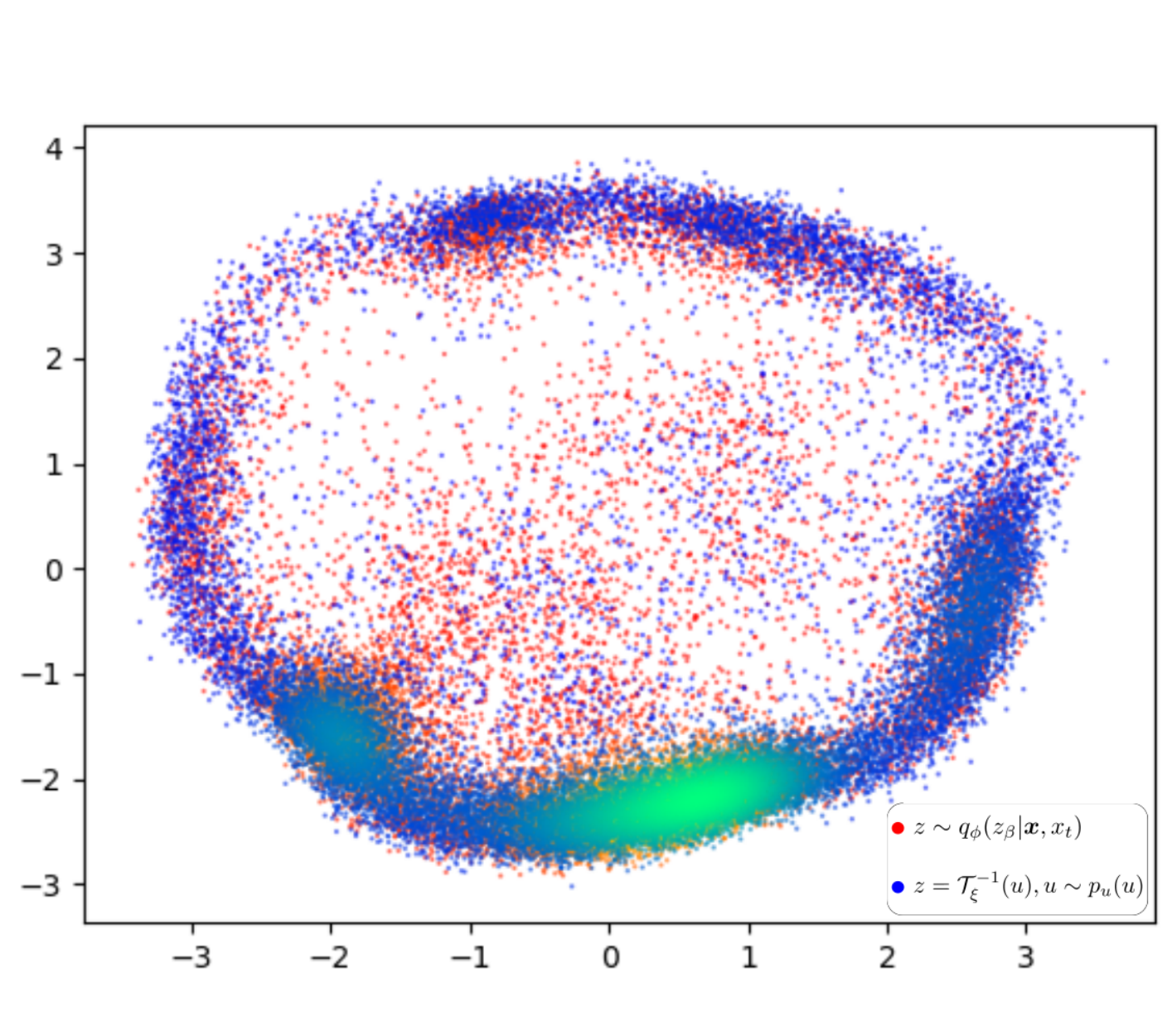}
    (b)
 \end{minipage}
\caption{2D-UMAP of prior and posterior samples without (a) and with (b) learning the normalizing flow transformation $\mathcal{T}_\xi$ for distribution alignment.}
\label{fig:sample_flow_prior}
\end{figure*}

\paragraph{Invertible Transformation $\boldsymbol{\mathcal{T}_\xi}$}
To highlight the need for learning an explicit mapping between the prior $p(z_\beta)$ and the posterior $q(z_\beta|\boldsymbol{x},x_t)$, we plot in Fig.~\ref{fig:sample_flow_prior} 2D UMAP~\cite{umap} visualizations of samples drawn from these distributions without and with using $\mathcal{T}_\xi$. Fig.~\ref{fig:sample_flow_prior} (a) shows a clear mismatch between both distributions. Fig.~\ref{fig:sample_flow_prior} (b) demonstrates that applying the transformation $\mathcal{T}_\xi$ helps to align prior and posterior, which is also reflected by the results discussed in the paragraph 'Behavior Sampling'. \\
Our normalizing flow model $\mathcal{T}_\xi$ is implemented as a stacked sequence of 15 invertible neural networks based on an input dimensionality of $D=1024$. Each consists of 3 blocks of subsequently applied actnorm \cite{glow}, affine coupling layers \cite{coupling_lay} and shuffling layers. The affine coupling layers consist of 2 fully connected layers with dimensionality $D=1024$. We trained the normalizing flow model on a single Titan Xp for 5 epochs with batchsize $64$ and ADAM~\cite{adam} optimizer with learning rate $6.5 \times 10^{-6}$.

\subsection{Protocols of Ablation Studies}
\label{sec:abl_protocoll}

\paragraph{Sample-Reality Classifier}
In Fig. 4 (b) of our main paper, we evaluate the quality of our generations with a recurrent binary classifier similar to \cite{mix-and-match}. The task of the classifier is to distinguish between 25k samples ground-truth sequences and 25k synthesized generations based on samples from the prior distribution. The classifier consists of a single layer GRU network with 256 hidden dimension for feature extraction, followed by a fully connected layer before applying the sigmoid function for binary classification. We optimize the classifier via stochastic gradient descent for 2k iterations, with a batch size of 256, a learning rate of 0.001 and a momentum of 0.9.

\paragraph{Average Regression Error (RE)}
In Tab. 1 of our main paper we provide an explicit quantitative evaluation of the disentanglement of posture and behavior. We adopt the experiments of \cite{mathieu16} and train a Multi-Layer Perceptron (MLP) consisting of 3 linear layers with 512, 256 and 51 neurons to predict the keypoint locations of postures in sequence $\boldsymbol{x}_\beta$ at different time-steps $T$ based on their corresponding extracted behavior representation $z_\beta$. Therefore, we train the MLP for 20 epochs with Adam \cite{adam} optimizer and a learning rate of $1 \times 10^{-3}$ on the test set as described in the main paper. Intuitively, if $z_{\beta}$ captures no information about posture, RE is high and converges to $0$ is lots of posture information is captured.

\paragraph{Action Classifier} In Sec. 4 of our main paper we evaluate the informativeness of the behavior representation $z_\beta$ by means of their benefit as a feature representation for action classification on Human3.6M dataset~\cite{h36m_pami} (\textit{acc.} values of the evaluated models in Tab. 1). For this purpose, we directly train a linear classifier on top of the frozen behavior encoder. For training and evaluation we use the same train-test split as described in the main paper. For the validation classifier which results in an test accuracy of $45\%$ ('gt:0.45', Tab. 1, main paper), we train a classifier with trainable feature representation which has the same architecture as our behavior encoder $q_\phi(z_\beta|\boldsymbol{x},x_t)$ to predict the action labels from ground truth sequences of 50 frames.

\subsection{Additional Results}
\label{sec:add_results}
Subsequently, we show additional visual results depicted as figures in this manuscript or as videos in the folder \textit{'videos'}.

\begin{figure*}[t]
    \centering
    \includegraphics[width=0.75\linewidth]{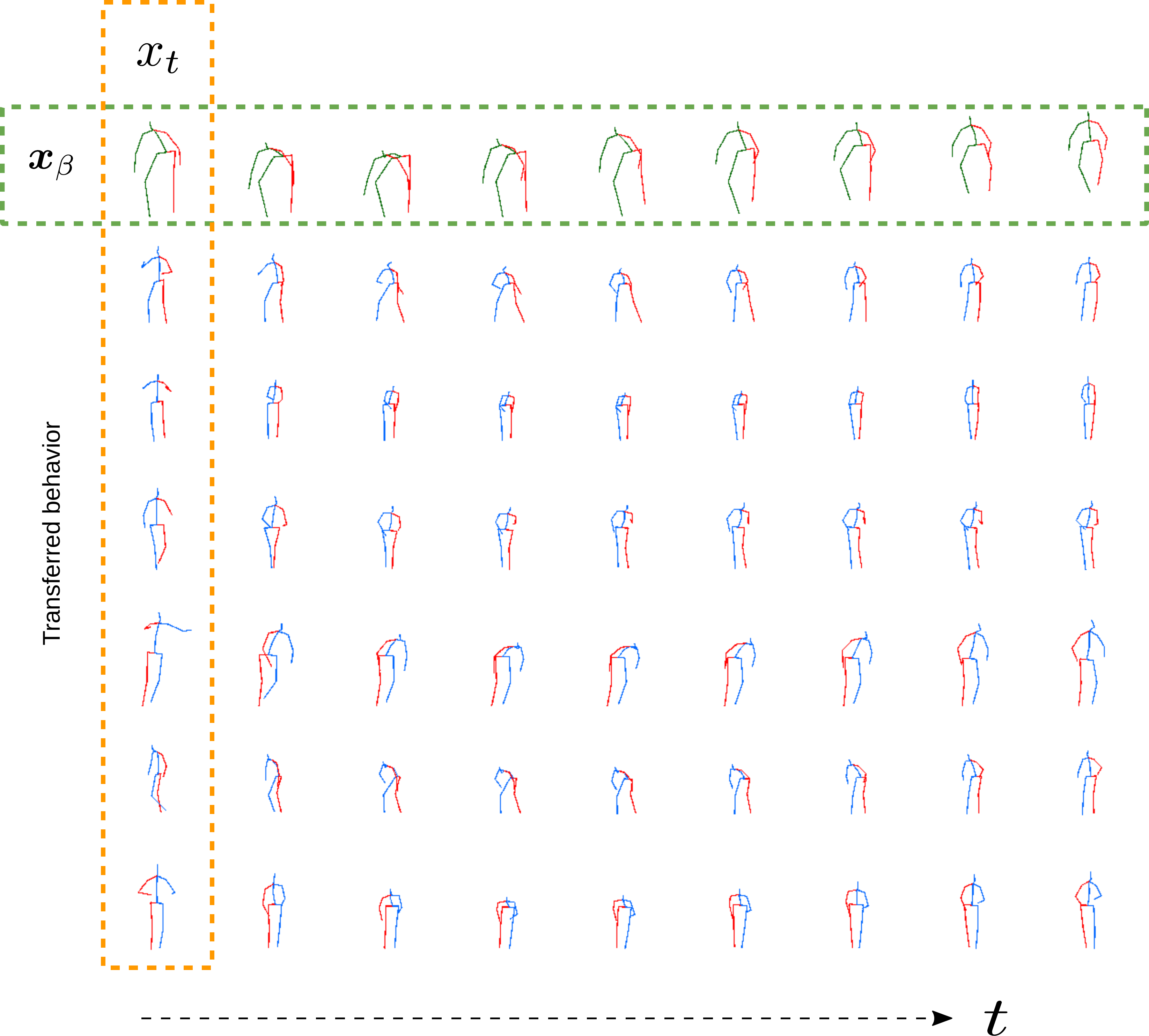}
    \caption{\textit{Behavior Transfer on Human3.6m \cite{h36m_pami}}. We transfer fine-grained, characteristic body dynamics of an observed behavior $x_{\beta}$ to unrelated, significantly different target postures $x_t$. Best viewed in PDF when zoomed in.}
    \label{fig:grid1}
\end{figure*}

\begin{figure*}[t]
    \centering
    \includegraphics[width=0.75\linewidth]{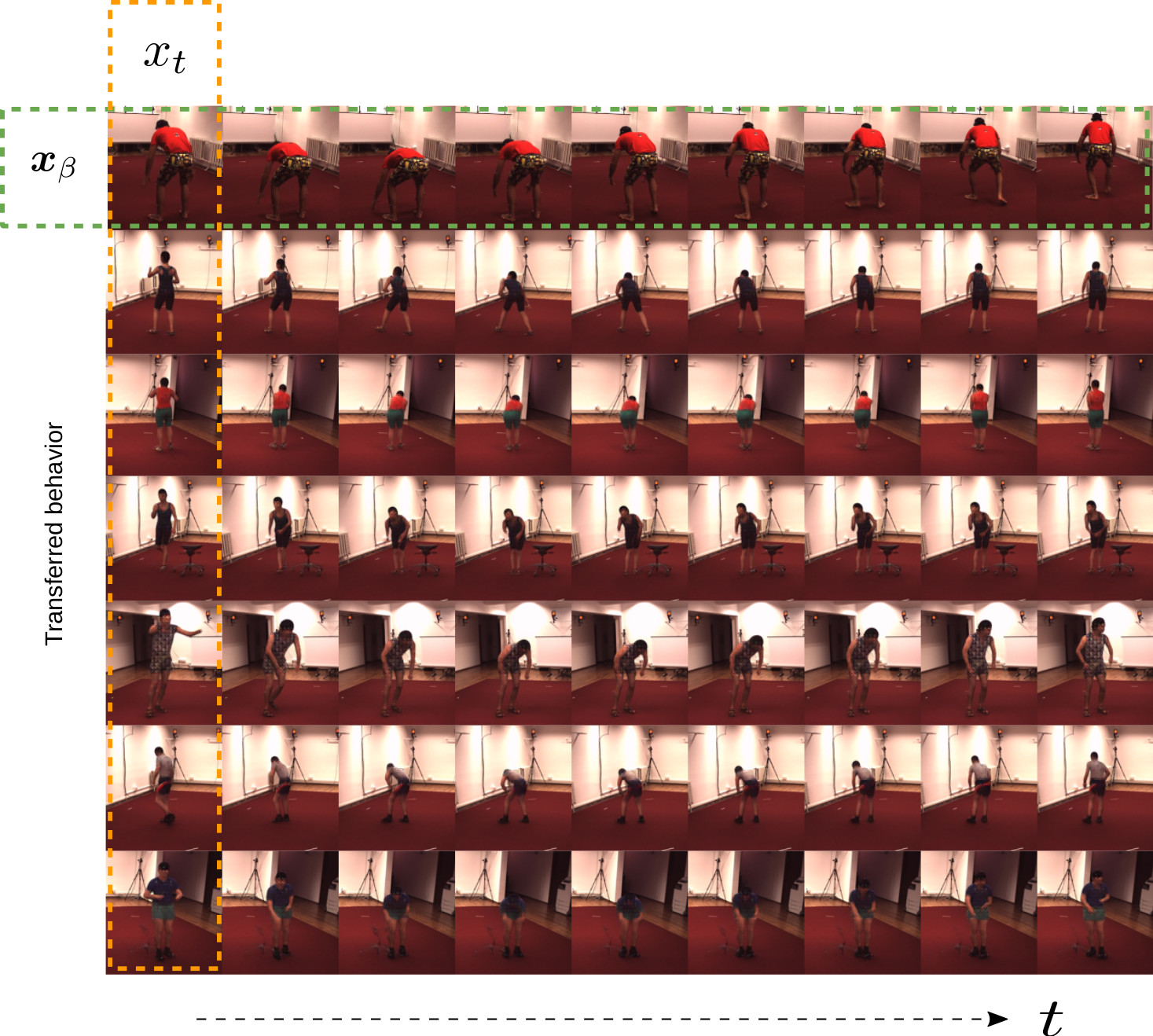}
    \caption{\textit{Translation of Fig.~\ref{fig:grid1} to RGB images}. We transfer fine-grained, characteristic body dynamics of an observed behavior $\boldsymbol{x}_{\beta}$ to unrelated, significantly different target postures $x_t$. Best viewed in PDF when zoomed in.}
    \label{fig:grid1_rgb}
\end{figure*}

\begin{figure*}[t]
    \centering
    \includegraphics[width=0.75\linewidth]{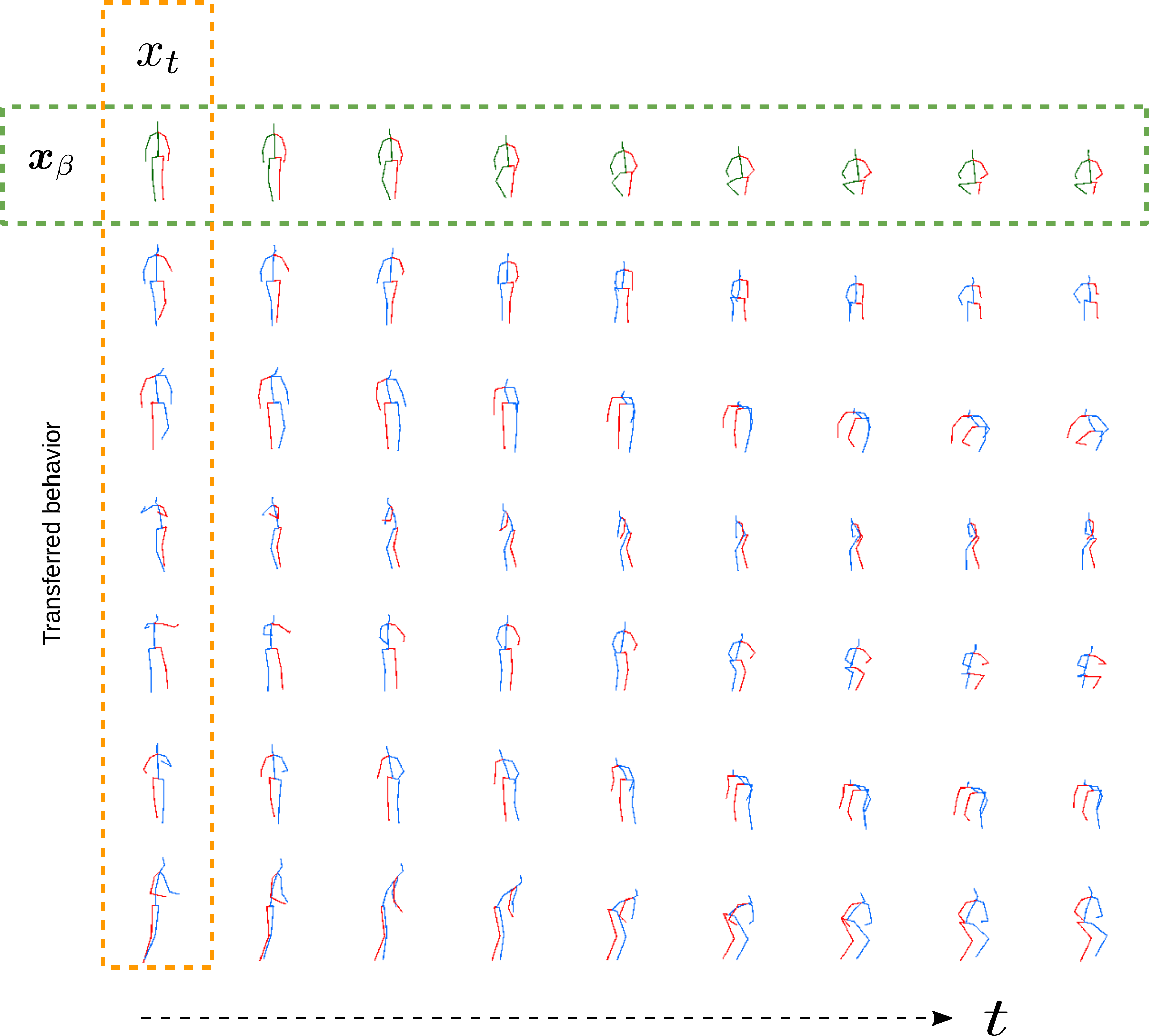}
    \caption{\textit{Behavior Transfer on Human3.6m \cite{h36m_pami}}. We transfer fine-grained, characteristic body dynamics of an observed behavior $x_{\beta}$ to unrelated, significantly different target postures $x_t$. Best viewed in PDF when zoomed in.}
    \label{fig:grid2}
\end{figure*}

\begin{figure*}[t]
    \centering
    \includegraphics[width=0.75\linewidth]{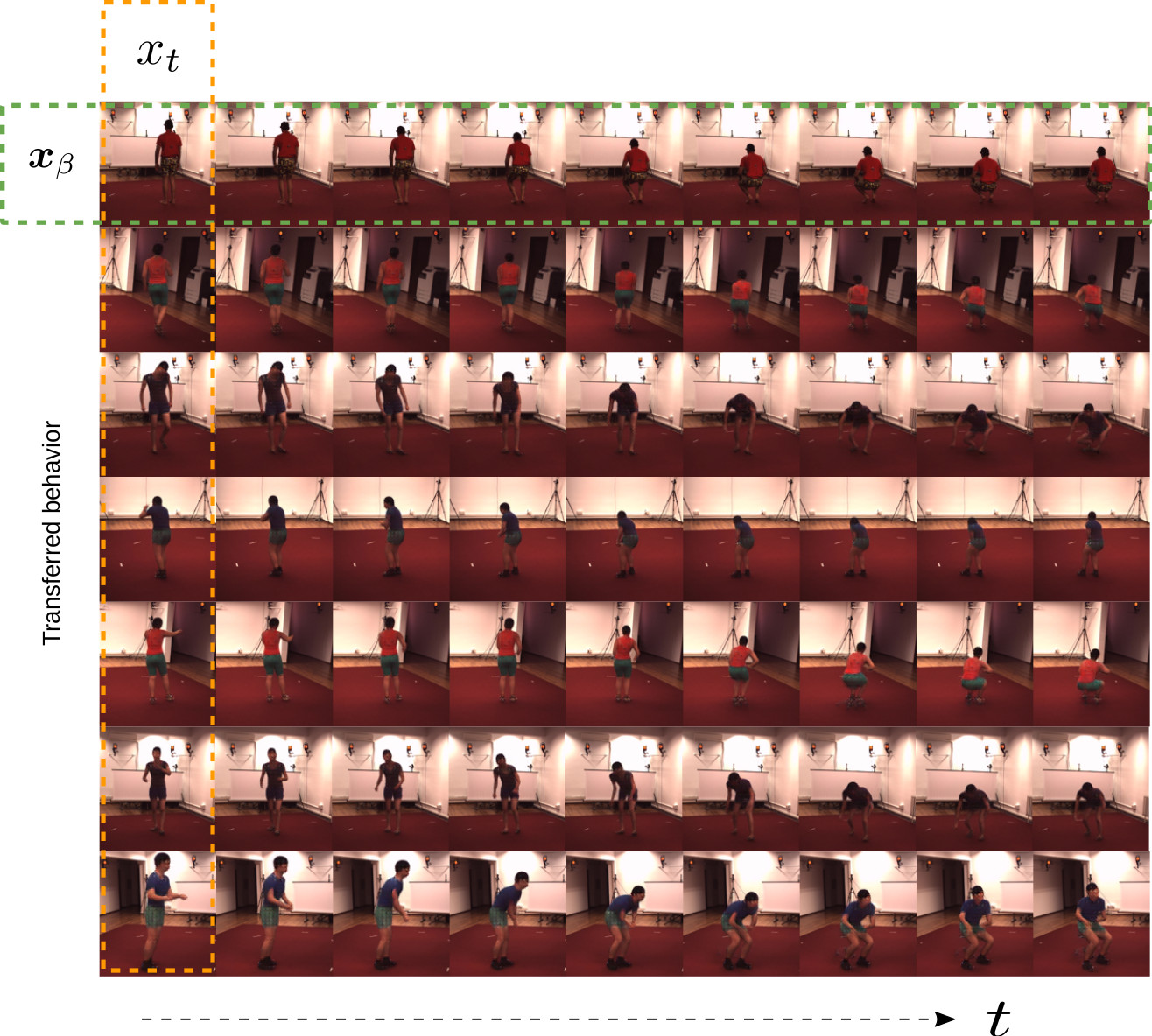}
    \caption{\textit{Translation of Fig.~\ref{fig:grid2} to RGB images}. We transfer fine-grained, characteristic body dynamics of an observed behavior $\boldsymbol{x}_{\beta}$ to unrelated, significantly different target postures $x_t$. Best viewed in PDF when zoomed in.}
    \label{fig:grid2_rgb}
\end{figure*}

\begin{figure*}[t]
    \centering
    \includegraphics[width=0.75\linewidth]{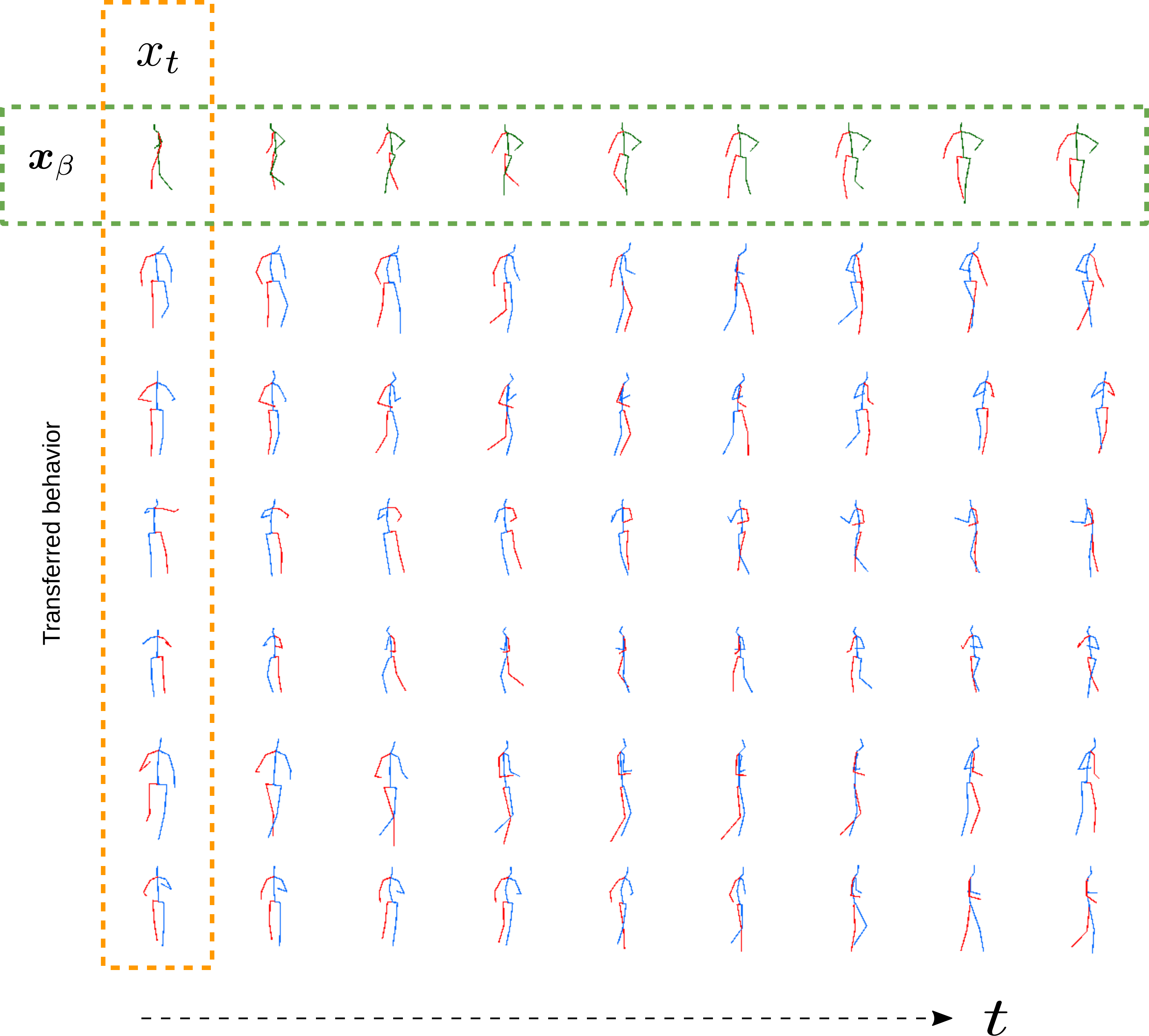}
    \caption{\textit{Behavior Transfer on Human3.6m \cite{h36m_pami}}. We transfer fine-grained, characteristic body dynamics of an observed behavior $x_{\beta}$ to unrelated, significantly different target postures $x_t$. Best viewed in PDF when zoomed in.}
    \label{fig:grid4}
\end{figure*}

\begin{figure*}[t]
    \centering
    \includegraphics[width=0.75\linewidth]{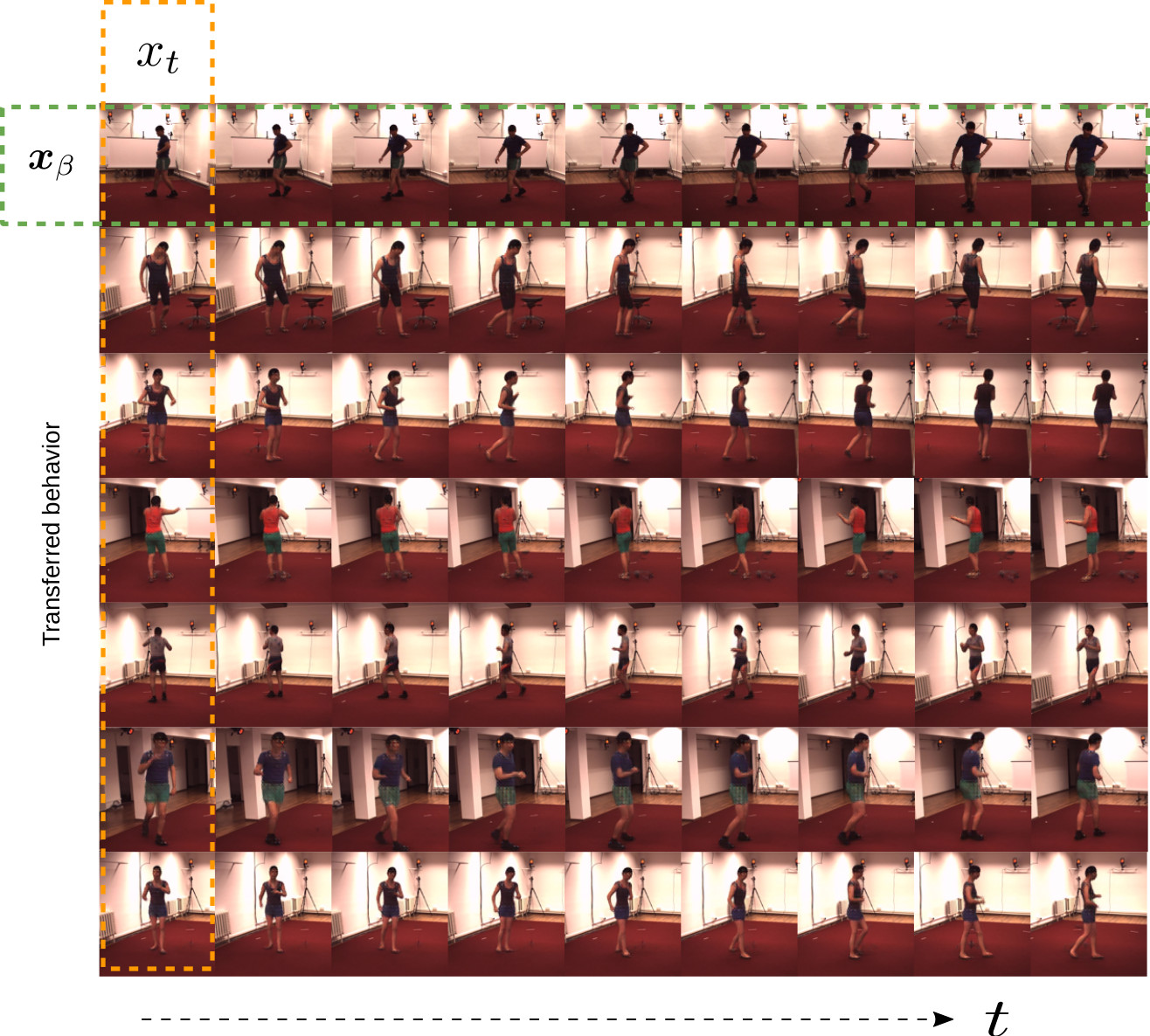}
    \caption{\textit{Translation of Fig.~\ref{fig:grid4} to RGB images}. We transfer fine-grained, characteristic body dynamics of an observed behavior $\boldsymbol{x}_{\beta}$ to unrelated, significantly different target postures $x_t$. Best viewed in PDF when zoomed in.}
    \label{fig:grid4_rgb}
\end{figure*}

\paragraph{Behavior Transfer}
We show more examples of behavior transfer in Fig.~\ref{fig:grid1}-\ref{fig:grid4_rgb}, both as postures and RGB images, similar to Fig. 3 of our main paper to further demonstrate the effectiveness of our approach.  Moreover, we also show videos based on both our model and the cAE/cVAE models which we quantitatively evaluated in Sec. 4 (Quantitative evaluation).
\\ \\
\textit{(i) cAE:} The video \textit{'behavior\_transfer\_CAE.mp4'} shows behavior re-enactments based on the cAE model. The topmost row depicts the source behavior sequence $\boldsymbol{x}_\beta$, while the leftmost column shows different target postures $x_t$. Based on these we show all pairwise combinations. We see that in general the cAE model quickly warps from $x_t$ to some early posture of $\boldsymbol{x}_\beta$. Next, it almost exactly copies the remaining posture sequence $\boldsymbol{x}_\beta$. Thus, given a certain $\boldsymbol{x}_\beta$ each re-enacted sequence is identical and independent of the given target pose $x_t$, rather than transferring only the behavior dynamics to the observed target postures. This is explained by the missing disentanglement of posture and behavior, which allows the cAE model to fully capture the complete posture information of $\boldsymbol{x}_\beta$ in $z_\beta$.
\\ \\
\textit{(ii) cVAE:} The video \textit{'behavior\_transfer\_CVAE.mp4'} shows behavior re-enactments based on the cVAE model. The topmost row depicts the source behavior sequence $\boldsymbol{x}_\beta$, while the leftmost column shows different target postures $x_t$. Based on these we show all pairwise combinations. We observe that this model predicts a likely future continuation based on the target posture $x_t$, thus not using the behavior representation $z_\beta$ for additional, dedicated information describing the source sequence $\boldsymbol{x}_\beta$. This is explained by posterior collapse, i.e. mean and variance of $q_\phi(z_\beta|\boldsymbol{x},x_t)$ collapsing to almost constant values.
\\ \\
\textit{(iii) Ours:}
The videos \textit{'behavior\_transfer1.mp4'} and \textit{'behavior\_transfer2.mp4'} show behavior re-enactments based on our proposed behavior transfer model. In both videos, the topmost row depicts the source behavior sequence $\boldsymbol{x}_\beta$, while the leftmost column shows different target postures $x_t$. Based on these we show all pairwise combinations. We see that our model extracts the behavior dynamics from diverse source sequences $\boldsymbol{x}_\beta$ and successfully transfers them to arbitrary target postures $x_t$ resulting in meaningful re-enactments of behavior $\beta$. Moreover, \textit{'behavior\_transfer1\_RGB.mp4'} and \textit{'behavior\_transfer2\_RGB.mp4'} show RGB video syntheses of our results using our model for posture-appearance transfer (see Appendix B and main paper).

\paragraph{Behavior Sampling}
\label{sec:sampling}
We now compare syntheses of novel behavior based on samples $z_\beta$ drawn from the prior distribution $p(z_\beta)$, with and without using the transformation $\mathcal{T}_\xi$ for correcting the mismatch with the posterior $q_\phi(z_\beta|\boldsymbol{x},x_t)$. For this purpose, we recursively synthesize behavior using sampled behavior representations $z_\beta$ and the last posture of the previously generated posture sequence. For detailed comparison, we show such a concatenated posture sequence without using $\mathcal{T}_\xi$ in video \textit{'sample\_loop\_prior.mp4'} and with $\mathcal{T}_\xi$ in \textit{'sample\_loop\_flow.mp4'}. We observe that the first suffers from synthesis artifacts due to out-of-distribution samples $z_\beta$, which in particularly become evident at the beginning of each behavior synthesis. In contrast, the recursively generated sequence using transformation $\mathcal{T}_\xi$ does not exhibit such artifacts and consequently results in a much smoother and more realistic sequence of diverse human behavior. 
\\
Moreover in video \textit{'samples.mp4'} we show behavior synthesis based on random sampling $z_\beta$ from the prior distribution which are then transformed using $\mathcal{T}_\xi$. The leftmost column depicts the target postures $x_t$ with each performing $6$ randomly sampled behaviors. Note, that for each target posture $x_t$ we use different samples $z_\beta$.

\paragraph{Behavior Nearest Neighbors}
To also demonstrate visually that our learned representation $z_\beta$ actually captures behavior dynamics while discarding posture information, we find nearest neighbours to the ground-truth training sequences. Therefore, we re-enact a source behavior $\boldsymbol{x}_\beta$ using a random target posture $x_t$. Next, we find its nearest neighbour in the training sequences based on \textit{(i)} distance between behavior representations $z_\beta$ and \textit{(ii)} average distances between postures sequences (based on alignment w.r.t. the pelvis keypoints). The video \textit{'nearest\_neighbors.mp4'} shows our results: Each column depicts a separate example showing the 'Source Behavior', the 'Nearest Neighbor based on Behavior representation', the 'Behavior Re-enactment of Source Behavior' and the 'Nearest Neighbor based on Posture', i.e. average posture distance. We observe that while there exist close training sequences in terms of posture, the nearest neighbors based on $z_\beta$ show \textit{similar} behavior dynamics while being \textit{dissimilar} in posture.

\paragraph{Behavior Interpolation}
To further analyze the regularity of our behavior representation $z_\beta$, we interpolate between the behavior observed in two sequences $\boldsymbol{x}_\beta^1$ and $\boldsymbol{x}_\beta^2$. To this end, we first extract their corresponding behavior representations $z_\beta^1, z_\beta^2$ and interpolate between them at equidistant steps, i.e. $(1 - \lambda) \cdot z_\beta^1 + \lambda \cdot z_\beta^2; \; \lambda \in \{0.0,0.2,0.4,0.6,0.8,1.0\}$. Next, we generate a sequence of interpolated behavior using our decoder $p_\theta(\boldsymbol{x}|z_\beta,x_t)$ with $x_t$ being the first frame of $\boldsymbol{x}_\beta^1$, respectively $\boldsymbol{x}_\beta^2$. Note, that for $\lambda \in \{0,1.0\}$ we basically reconstruct the source sequences $\boldsymbol{x}_\beta^1$, $\boldsymbol{x}_\beta^2$. We show the resulting posture sequences in \textit{'interpolations\_01.mp4'}-\textit{'interpolations\_03.mp4'} and with additional RGB image overlay in \textit{'interpolations\_rgb\_01.mp4'}-\textit{'interpolations\_rgb\_03.mp4'}.

\paragraph{Behavior Generalization}
We now demonstrate the robustness of our proposed model to unseen behavior dynamics by leaving out sets of entire classes during training\footnote{Note, that we only use labels for excluding training sequences in this experiment, but not for the training procedure itself.} and, subsequently, performing behavior transfers based on source sequences $\boldsymbol{x}_\beta$ sampled from these classes. \\
We show results for both excluding walking actions ('walking', 'walking dog', 'walking together') in video \textit{'behavior\_transfer\_generalization\_walking.mp4'} and sitting actions ('sitting', 'sitting down', 'purchases') in video \textit{'behavior\_transfer\_generalization\_sitting.mp4'}. The top rows depict the source behaviors $\boldsymbol{x}_\beta$ and the leftmost columns show the target postures $x_t$. In both cases our model is able to correctly infer the body dynamics characterizing these actions.

\section{Posture and Appearance Model}
\label{sec:shape_appearance}
Our proposed conditional framework for disentanglement can also be applied for the task of appearance transfer. Instead of disentangling posture from behavior, we disentangle posture from appearance of persons depicted on static images and use the resulting model to generate RGB video sequences based on the re-enacted posture sequences as reported in the main paper. Note that the posture and appearance model operates on 2D keypoints. Therefore, we project the 3D keypoints locations of the re-enacted sequences onto the image plane. Subsequently, we provide implementation details and additional experiments on DeepFashion \cite{deepfashion_data} and Market1501 \cite{mark_data} datasets.

\subsection{Architecture and Losses}
Our model for appearance transfer is based on a UNet architecture similar to VUnet~\cite{vunet}. The UNet maps from posture $x_t$, i.e. keypoint skeletons, to RGB images with appearance information added at the bottleneck which is extracted from some image $I_\alpha$ by an appearance encoder. Now, we provide implementation details for the posture- and appearance encoder, as well as the decoder.


\paragraph{Appearance encoder:} The appearance encoder, which is the equivalent of the behavior encoder for the task of posture-appearance disentanglement, is implemented as a fully convolutional network. We gradually downsample the input image $I_\alpha$ up to a spatial size of $4\times4$. Each downsampling stage consists of 2 ResNet blocks and downsampling is performed using a convolutional layer with stride $2$. We double the number of feature channels at every stage up to a maximum number of $128$ which is then kept fixed. At the bottleneck we compute mean and variance both based on the layer outputs of spatial size $8$ and $4$~\cite{vunet}.

\paragraph{UNet encoder and decoder}: Both the encoder and decoder branch of the UNet are similarly designed as the appearance encoder with skip connections connecting them at each stage. For upsampling in the decoder we use bilinear interpolation. At the bottleneck, we concatenate the feature maps of the posture stream with the encodings of the appearance encoder.

\paragraph{Auxilliary decoder:} 
The auxilliary decoder consists of two convolutional layers with kernel size $8$ and $4$. It takes the appearance encodings as input (both at spatial sizes 8 and 4) and outputs one vector for each with dimensionality $256$. Following that, we add $6$ linear layers (with dimenionalities 512,512,256,128,64,34) to predict the posture keypoints. 

\paragraph{Optimization:}
For optimizing the likelihood $\mathbb{E}_{q_\phi}(\boldsymbol{x}|z_\alpha,x_t)$ similar to Eq. (5), with $z_\alpha$ denoting the appearance encoding, we employ both standard pixel-wise mean squared error and a perceptual loss~\cite{vunet}. The latter is a feature matching loss and often used to emphasize on structural information such as contours and texture. It is formulated as
\begin{align}
    \mathcal{L}_{\alpha,feat} = \sum_k \lambda_k \cdot \lVert F_k(I_\alpha) - F_k(\tilde{I}_\alpha) \rVert_1 \;
\end{align}
where $F_k$ denote feature layers of a pretrained VGG19 network \cite{vgg}, the weights $\lambda_k$ control the amount contribution of each layer $k$, $I_\alpha$ is the target image to be reconstructed and $\tilde{I}_\alpha$ its reconstruction, i.e. output of the decoder. Note that the model does not require image pairs of persons with the same appearance label and can hence be trained solely by reconstructing static image frames.

\begin{table}
    \centering
    \begin{tabular}{l | cc | cc }   
    \toprule    
    \multirow{2}{*}{Method} & \multicolumn{2}{c}{DeepFashion}& \multicolumn{2}{c}{Market1501}\\
    ~ & IS & SSIM & IS & SSIM \\
    \midrule
    \hline
    VUnet (Esser et al. 2018) & 3.09 & 0.79  & 3.21 & \textbf{0.35}  \\
    DIG \cite{ma2017disentangled}  & \textbf{3.23} & 0.61  & 3.44 & 0.10  \\
    PG$^2$ \cite{app_shape_dis_1}  & 3.09 & 0.76  & \textbf{3.46} & 0.25 \\
    Ours & 3.08 & \textbf{0.80} & 3.16 & \textbf{0.35}  \\
    \end{tabular}
     \caption{Evaluation of our shape-appearance transfer model based on image quality metrics on DeepFashion~\cite{deepfashion_data} and Market1501~\cite{mark_data} (Reconstruction Setting).}
    \label{fig:shapeAppTransfer_eval}
\end{table}

\subsection{Training Details}

\paragraph{Human3.6m}
On Human3.6M, we train our appearance model for 150k iterations using ADAM optimizer~\cite{adam} with learning rate $0.0005$. During the alternating optimization, we perform $5$ update steps of the auxiliary decoder for each update step of the appearance model. Further, we set $I_{\text{KL}}=1000$, $\gamma_\text{C} = 1, \lambda_k = 1 \, \forall \, k$ and use no inplane normalization~\cite{vunet}.

\paragraph{DeepFashion}
On DeepFashion~\cite{deepfashion_data}, we train our appearance model for 200k iterations using ADAM optimizer~\cite{adam} with learning rate $0.0005$. During the alternating optimization, we perform $10$ update steps of the auxiliary decoder for each update step of the appearance model. Further, we set $I_{\text{KL}}=1000$, $\gamma_\text{C} = 5, \lambda_k = 1 \, \forall \, k$ and use inplane normalization~\cite{vunet}.

\paragraph{Market}
On Market~\cite{mark_data}, we train our appearance model for 150k iterations using ADAM optimizer~\cite{adam} with learning rate $0.0005$. During the alternating optimization, we perform $5$ update steps of the auxiliary decoder for each update step of the appearance model. Further, we set $I_{\text{KL}}=1000$, $\gamma_\text{C} = 1, \lambda_k = 1 \, \forall \, k$ and use inplane normalization~\cite{vunet}.

\subsection{Additional Results}
\begin{figure*}[t]
    \centering
    \includegraphics[width=\linewidth]{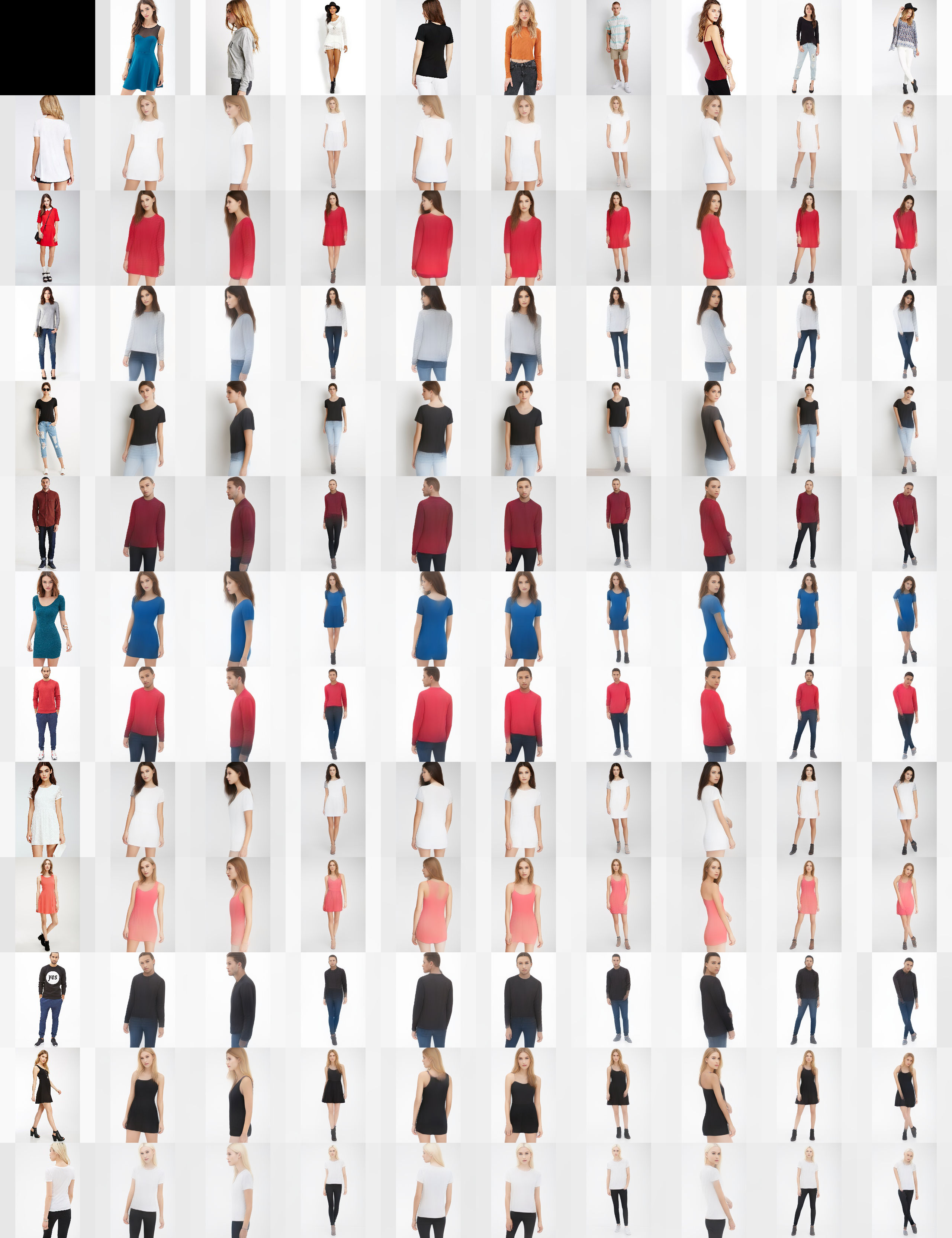}
    \caption{Posture-Appearance transfer on \textit{DeepFashion}~\cite{deepfashion_data}. Top row depicts target posture and leftmost row depicts source appearance.}
    \label{fig:transfer_deepfashion}
\end{figure*}

\begin{figure*}[t]
    \centering
    \includegraphics[width=0.95\linewidth]{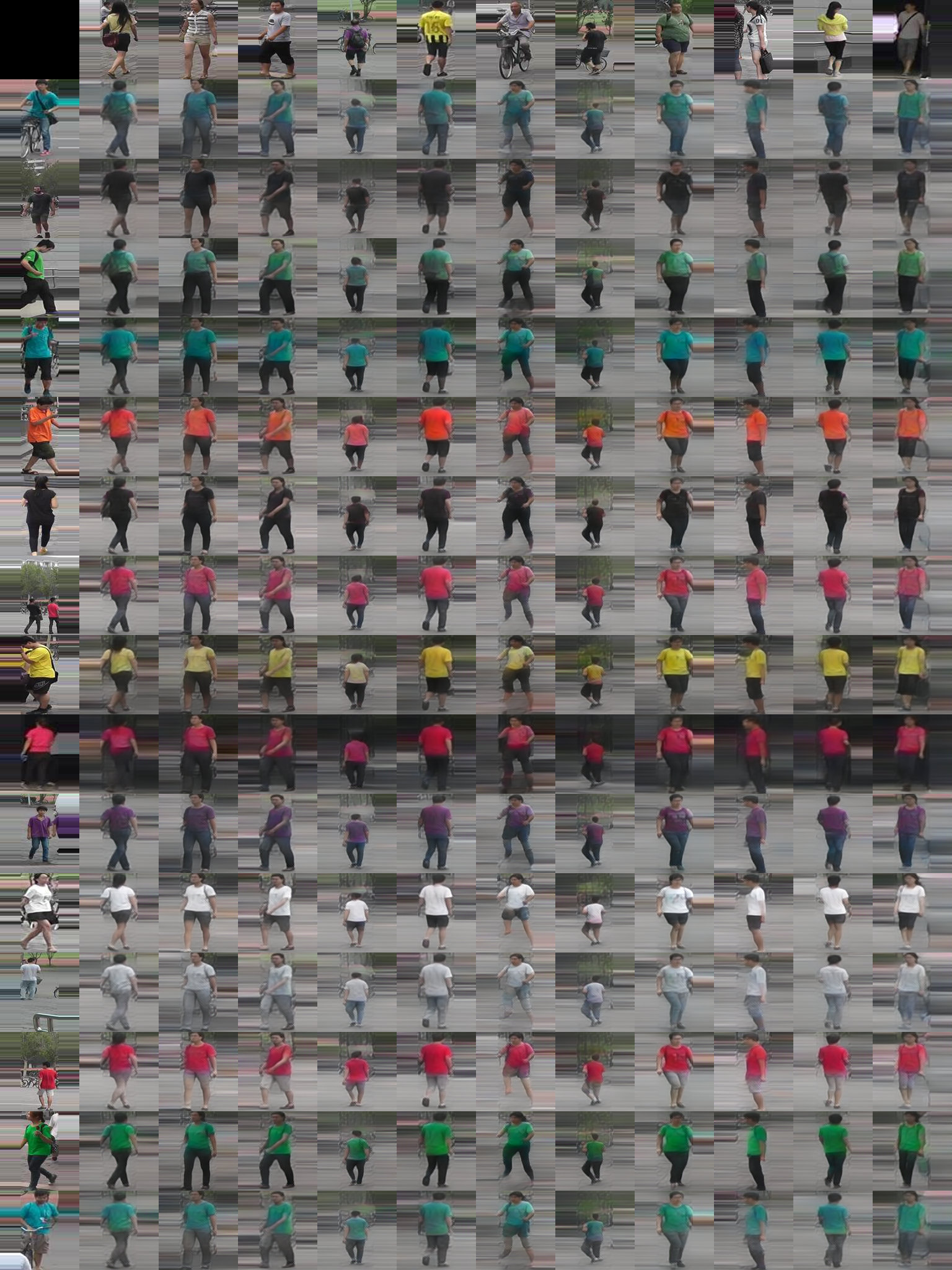}
    \caption{Posture-Appearance transfer on \textit{Market1501}~\cite{mark_data}. Top row depicts target posture and leftmost row depicts source appearance.}
    \label{fig:transfer_market}
\end{figure*}

To evaluate our model for appearance transfer also on established datasets dedicated to this task, we report in Fig.~\ref{fig:shapeAppTransfer_eval} Inception Score (IS)~\cite{inception_score} and Structured Similarity (SSIM)~\cite{ssim} on DeepFashion~\cite{deepfashion_data} and Market1501~\cite{mark_data} dataset. We observe that our model performs competitively with the state-of-the-art on human shape-appearance transfer, thus indicating the general applicability of our disentanglement framework. Moreover, in Fig.~\ref{fig:transfer_deepfashion} we provide example appearance transfers. Top rows depict the target posture and leftmost columns depict the source appearance. Similarly, in Fig.~\ref{fig:transfer_market} we show transfers between posture and appearance for the Market1501~\cite{mark_data} dataset.


\end{document}